\documentclass[twoside]{article}

\relax

\newlength\titlebox 
\usepackage[round]{natbib}

\usepackage{microtype}
\usepackage{graphicx}
\usepackage{booktabs} 
\usepackage{xcolor}
\usepackage{wrapfig}
\usepackage{comment}
\usepackage{bm}
\usepackage{caption}
\usepackage{subcaption}
\usepackage{dsfont}
\newcommand{\mathbbm}[1]{\mathds{#1}}

\definecolor{mydarkred}{rgb}{0.6,0,0}
\definecolor{mydarkgreen}{rgb}{0,0.6,0}
\usepackage[colorlinks,
linkcolor=mydarkred,
citecolor=mydarkgreen]{hyperref}

\usepackage{amsmath}
\usepackage{amssymb}
\usepackage{mathtools}
\usepackage{amsthm}

\usepackage{tikz}
\usetikzlibrary{shapes,decorations,arrows,calc,arrows.meta,fit,positioning}
\tikzset{
    -Latex,auto,node distance =1 cm and 1 cm,semithick,
    state/.style ={circle, draw, minimum width = 0.7 cm},
    point/.style = {circle, draw, inner sep=0.04cm,fill,node contents={}},
    bidirected/.style={Latex-Latex,dashed},
    el/.style = {inner sep=2pt, align=left, sloped}
}

\usepackage[capitalize,noabbrev]{cleveref}

\theoremstyle{plain}
\newtheorem{thm}{Theorem}[section]
\newtheorem{prop}[thm]{Proposition}
\newtheorem{lem}[thm]{Lemma}
\theoremstyle{definition}
\newtheorem{assum}[thm]{Assumption}
\theoremstyle{remark}

\crefname{assum}{Assumption}{Assumptions}

\usepackage{./style/mydef}

\global\long\def\calY{\mathcal{Y}}%
\global\long\def\calF{\mathcal{H}}%

\global\long\def\calF{\mathcal{F}}%

\global\long\def\inner#1#2{\left\langle #1,#2\right\rangle }%

\global\long\def\la{\langle}%

\global\long\def\ra{\rangle}%

\global\long\def\inner#1#2{\left\langle #1,#2\right\rangle }%

\global\long\def\la{\langle}%

\global\long\def\ra{\rangle}%

\global\long\def\given{\vert}%

\newcommand{\citetMastouri}[1][]{\citet[#1]{Mastouri2021Proximal}~}
\begin{document}

%

%

\title{Kernel Single Proxy Control for Deterministic Confounding}

\author{ Liyuan Xu\\ Gatsby Computational Neuroscience Unit \and Arthur Gretton \\ Gatsby Computational Neuroscience Unit} 
\date{}

\maketitle

\begin{abstract}
  We consider the problem of causal effect estimation with an unobserved confounder, where we observe a \emph{single} proxy variable that is associated with the confounder. Although it has been shown that the recovery of an average causal effect is impossible in general from a single proxy variable, we show that causal recovery is possible if the outcome is generated deterministically. This generalizes existing work on causal methods with a single proxy variable to the continuous treatment setting.
We propose two kernel-based methods for this setting: the first based on the two-stage regression approach, and the second based on a maximum moment restriction approach. We prove that both approaches can consistently estimate the causal effect, and we empirically demonstrate that we can successfully recover the causal effect on challenging synthetic benchmarks.
\end{abstract}

\section{Introduction}
Causal learning involves estimating the impact of our actions on the world. For instance, we might want to understand how flight ticket prices affect sales, or how a specific medication increases the chance of recovery. The action we take is referred to as a ``treatment,'' which leads to a specific ``outcome.'' Determining the causal effect solely from observational data is often challenging, however, because the observed relationship between treatment and outcome can be influenced by an unobserved factor known as a ``confounder.'' In the case of the medical example, patient health might act as a confounder: if the patient is weaker, then he or she might require a higher dose of medicine, and yet have a lower chance of survival.
Thus, from observing historical data,  the dose may be \emph{negatively correlated} with the chance of recovery; an instance of  \emph{Simpson's paradox}.
Hence, it is necessary to address the bias caused by the confounder to assess the true effect.

Although many methods for causal inference methods assume that there are \emph{no unmeasured confounders}, this can be too restrictive, and it is often difficult to determine how the confounder affects treatment assignments and outcomes. A milder assumption is that we have access to {\it proxy variables}, which are correlated to the confounder. It is known that we cannot in general recover the true causal effect with a single proxy variable \citep{pearl10bias,Kuroki2014Mesurement}: for this reason, studies
have focused on the case where two proxy variables are available \citep{Kuroki2014Mesurement,Miao2018Identifying,Deaner2018Proxy,Mastouri2021Proximal,Rahul2020KernelProxy,xu2021deep}.


In this paper, we introduce a novel sufficient condition which characterizes when causal effect estimation from a {\em single} proxy variable is possible. Our key assumption is \emph{deterministic confounding}, meaning that the outcome is deterministically generated given the treatment and the confounder. This generalizes \emph{Single Proxy Control} \citep{Tchetgen2023Single}, which considers the setting where the confounder is the potential outcome \citep{Rubin2005}. As we will see, the deterministic confounding assumption can be also be understood with reference to a similar requirement for nonlinear independent component analysis \citep{hyvarinen19a}. 

Given the deterministic confounding assumption, we show that we can use the outcome itself as a ``proxy variable'' in the setting of Proxy Causal Learning (PCL)  \citep{Deaner2018Proxy,Miao2018Identifying}. This insight enables us to use a technique adapted from kernel PCL  \citep{Rahul2020KernelProxy,Mastouri2021Proximal} to estimate the causal effect. We further conduct a theoretical and empirical sensitivity analysis on the violation of the deterministic confounding assumption. We also derive a novel closed-form solution for kernel PCL methods with improved numerical stability, which may be of independent interest. 

This paper is structured as follows. After reviewing the related work, we present an overview of the problem setting, and of relevant background material on kernel methods in \cref{sec:preliminary}. In \cref{sec:identification}, we establish conditions under which the causal effect can be recovered from the observed variables. Then, in \cref{sec:method}, we introduce our estimation procedures: namely, the two-stage least squares and moment matching approaches, as in \citetMastouri, with consistency results for each. In \cref{sec:experiment}, we report the empirical performance of the proposed method on synthetic causal data. We also explore the case where the deterministic confounding assumption is violated, and show that our proposed methods can still obtain a reasonable estimate of the causal relationship.

\paragraph{Related Work} 

Despite the negative results shown by \citet{Kuroki2014Mesurement}, various attempts have been made to estimate causal effects from proxy variables. CEVAE \citep{Louizos2017Causal} uses a VAE \citep{Kingma2014VAE} to recover the distribution of confounders using a single proxy variable. Although it lacks theoretical justification and can be unstable \citep{rissanen2021a}, CEVAE has shown strong empirical performance in several settings.

\emph{Proxy causal learning (PCL)} \citep{Deaner2018Proxy,Miao2018Identifying} guarantees to recover the true causal effect given \emph{two} of proxy variables, under existence and identifiability conditions. The original work of \citet{Miao2018Identifying} provides a closed-form solution in the case of categorical treatments and outcomes, and derives a Fredholm equation of the first kind to characterize the continuous case. Solutions for the latter case have been obtained where the causal relationships are modeled using a dictionary of smooth basis functions \citep{Deaner2018Proxy}, RKHS functions \citep{Mastouri2021Proximal,Rahul2020KernelProxy}, and neural networks \citep{xu2021deep,kompa2022deep,Kallus2021Causal}. Our main contribution is to show that we can use the outcome as a proxy in PCL given an additional deterministic assumption.

\emph{Negative control outcome} only requires \emph{one} proxy variable, but  places stricter assumptions on the confounding structure.  \citet{Tchetgen2023Single} introduced \emph{Single Proxy Control}, which generalizes the \emph{Difference-in-Difference (DiD)} method \citep{angrist_mostly_2008,did}. Although the work relaxes the strict parallel trending assumption in DiD, they consider only the case of binary treatment, and require the proxy to reveal the potential outcome \citep{Rubin2005} of the treated. Our method generalizes Single Proxy Control to continuous treatments and a more general confounding mechanism. A relation with nonlinear independent component analysis (NICA) \citep{hyvarinen19a} is described in the next section, once the required notation has been introduced.

\section{Preliminaries} \label{sec:preliminary}

\paragraph{Notation.} Throughout the paper, a capital letter (e.g. $A$) denotes a random variable, and we denote the set where a random variable takes values by the corresponding calligraphic letter (e.g. $\mathcal{A}$). 
The symbol $\prob{\cdot}$ denotes the probability distribution of a random variable (e.g. $\prob{A}$). We use a lowercase letter to denote a realization of a random variable (e.g. $a$). We denote the expectation over a random variable as $\mathbb{E}[\cdot]$ and $\|f\|_{\prob{\cdot}}$ as the $L^2$-norm of a function $f$ with respect to $\prob{\cdot}$; i.e. $\|f\|_{\prob{A}} = \sqrt{\expect[A]{f^2(A)}}$.

\subsection{Problem Setting} \label{sec:problem-settings}

\begin{figure*}
    \centering
    \begin{subfigure}{0.22\linewidth}
    \centering
    \begin{tikzpicture}
        \node[state] (eps) at (0, 0) {$U$};
        \node[state, fill=yellow] (x) at (-1.,-1.5) {$A$};
    
        \node[state, fill=yellow] (y) at (1., -1.5) {$Y$};
        \node[state, fill=yellow] (w) at (1.5, 0) {$W$};
        
        \path[very thick] (x) edge (y);
        \path[very thick] (eps) edge (y);
        \path (eps) edge (x);
        \path[bidirected] (eps) edge (w);
    \end{tikzpicture}    
    \subcaption{Our setting}
    \label{fig:causal-graph}
    \end{subfigure}
     \begin{subfigure}{0.23\linewidth}
     \centering
    \begin{tikzpicture}
        \node[state] (eps) at (0, 0) {$U$};
        \node[state, fill=yellow] (x) at (-1.,-1.5) {$A$};
    
        \node[state, fill=yellow] (y) at (1., -1.5) {$Y$};
        \node[state, fill=yellow] (w) at (1.5, 0) {$W$};
        \node[state, fill=yellow] (z) at (-1.5, 0) {$Z$};
        \path (x) edge (y);
        \path (eps) edge (y);
        \path (eps) edge (x);
        \path (eps) edge (w);
        \path[bidirected] (eps) edge (z);
        \path (z) edge (x);
        \path[bidirected] (w) edge (y);
    \end{tikzpicture}    
    \subcaption{PCL setting}
    \label{fig:pcl-causal-graph}
    \end{subfigure}
    \begin{subfigure}{0.28\linewidth}
    \begin{tikzpicture}
        \node[state] (eps) at (1., 0) {$Y^{(1)}$};
        \node[state] (eps2) at (-1., 0) {$Y^{(0)}$};
        \node[state, fill=yellow] (x) at (-2.,-1.5) {$A$};
    
        \node[state, fill=yellow] (y) at (0, -1.5) {$Y$};
        \node[state, fill=yellow] (w) at (2, -1.5) {$W$};
        
        \path[very thick] (x) edge (y);
        \path[very thick] (eps) edge (y);
        \path[very thick] (eps2) edge (y);
        \path (eps) edge (x);
        \path (eps2) edge (x);
        \path[bidirected] (eps) edge (w);
        \path[bidirected] (eps) edge (eps2);
    \end{tikzpicture}    
    \caption{Single Proxy Control setting}
    \label{fig:single-proxy-causal-graph}
    \end{subfigure}
    \begin{subfigure}{0.23\linewidth}
    \centering
    \begin{tikzpicture}
        \node[state] (eps) at (0., 0.3) {$U$};
        
        \node[state, fill=yellow] (y) at (0, -1.5) {$Y$};
        \node (f) at (-0.5, -0.75) {$g_a(\cdot)$};
        \node[state, fill=yellow] (w) at (1., -0.75) {$W$};
        
        \path[very thick] (eps) edge (y);
        \path[bidirected] (eps) edge (w);
    \end{tikzpicture}    
    \caption{Nonlinear ICA setting}
    \label{fig:ica-causal-graph}
    \end{subfigure}
    \caption{Causal graphs with proxy variables.  Here, the bidirectional arrows mean that we allow an arrow in either direction, or even a common ancestor variable; and the thick arrow indicates a deterministic relationship between the variables. }
    
\end{figure*}

We begin by describing the problem setting. We observe the treatment $A\in\mathcal{A}$ and corresponding outcome $Y \in \mathcal{Y}$. We assume that there exists an unobserved confounder $U \in \mathcal{U}$ that affects both $A$ and $Y$. Our goal is to estimate the structural function $f_\mathrm{struct}$ defined as 
\begin{align*}
    f_\mathrm{struct}(\tilde a)  =  \expect[U]{\expect[Y]{Y|A=\tilde a, U}},
\end{align*}
for a given test point $\tilde a \in \mathcal{A}$. This function is also known as the \textit{Dose-Response Curve} or \textit{Average Potential Outcome}. The challenge in estimating $f_\mathrm{struct}$ is that the confounder $U$ is not observable ---  we cannot estimate the structural function from observations $A$ and $Y$ alone. To deal with this, we assume access to a proxy $W$, which satisfies the following two assumptions:
\begin{assum} \label{assum:structural}
We assume $W \indepe (A,Y) | U$.
\end{assum}
\begin{assum} \label{assum:completeness-confounder}
For any square integrable function $l: \mathcal{U} \to \mathbb{R}$, the following conditions hold for all $a \in \mathcal{A}$:
\begin{align*}
    &\expect{l(U) \mid A= a, W=w}=0\quad \forall w \in \mathcal{W} \\
    &\quad \Leftrightarrow \quad l(u) = 0 \quad \prob{U}\text{-}\mathrm{a.e.}
\end{align*}
\end{assum}

Figure~\ref{fig:causal-graph} shows the causal graph describing these relationships. We may additionally consider an observable confounder, which is discussed in \cref{sec:observable-confounder}.

The assumptions on the proxy $W$ are similar to those made in \emph{Proxy Causal Learning (PCL)} \citep{Deaner2018Proxy,Miao2018Identifying,Mastouri2021Proximal,Rahul2020KernelProxy,xu2021deep,Kallus2021Causal}. Our setting only assumes access to a single proxy $W$, however, while PCL requires an additional treatment-proxy $Z$ that satisfies $Z \indepe Y | A, U$ as in \cref{fig:pcl-causal-graph}. Instead of observing this proxy, we place deterministic confounding assumptions on the outcome $Y$, as follows.

\begin{assum} \label{assum:deterministic-confounding}
    There exists a \emph{deterministic} function $\gamma_0(a,u)$ such that $Y = \gamma_0(A,U)$ and $\|\gamma_0(a,U)\|_{\prob{U}} \leq \infty$ for all $a\in\calA$.
\end{assum}

This deterministic confounding assumption is unique in our setting: we will further discuss its necessity in \cref{sec:identification}. Our main contribution is to show that we may use outcome $Y$ as the treatment-proxy $Z$ in PCL if we make \cref{assum:deterministic-confounding}. Further details can be found in \cref{sec:connection-to-PCL}.

\paragraph{Applications} One typical application is a time-evolving deterministic process. For example, in a chemical synthesis plant, the amount of the final product $Y$ can be computed given  perfect knowledge of the plant state $U$ and the control signal $A$, since the reaction process can be modeled by a deterministic differential equation \citep{ChemGymRL}.

Single Proxy Control \citep{Tchetgen2023Single}, whose causal graph is shown in \cref{fig:single-proxy-causal-graph}, also considers deterministic confounding. From the definition of the potential outcome \citep{Rubin2005},  \cref{assum:deterministic-confounding} holds with $\gamma_0(a,Y^{(0)},Y^{(1)}) = \mathbbm{1}(a=0)Y^{(0)} + \mathbbm{1}(a=1)Y^{(1)}$. Single Proxy Control applies only to the binary treatment case, however, whereas our setting generalizes the approach to continuous treatments.

A related setting is nonlinear independent component analysis (NICA) \citep{hyvarinen19a}, in which  the observation $Y$ is given by a {\em smooth, deterministic, invertible}, nonlinear function of unobserved latents $U$, as shown in \cref{fig:ica-causal-graph}. Denoting this function as $g$, \citet{hyvarinen19a} prove that $g$ can be recovered up to certain indeterminacies if  we have an informative proxy variable $W$, and assume that the entries of $U$ are mutually independent given $W$. While we share our {\em deterministic confounding} assumption and use of an informative proxy with NICA,  our goal is different: we do not  recover $U$, nor do we require assumptions concerning the dependence structure within $U$. Our aim is rather to estimate the causal effect $\expect{g_a(U)}$ under intervention  $a$ and confounding $U$, where we have modeled a {\em family} of functions $g_a$ indexed by $a$. 

 
\subsection{Reproducing Kernel Hilbert Space}
For any space $\mathcal{F} \in \{\calA, \calY, \calW\}$, let $k_\calF: \mathcal{F}\times\mathcal{F}\to\mathbb{R}$ be a positive semidefinite kernel. Here, we assume all kernel functions are characteristic \cite{SriGreFukLanetal10} and bounded, $\max_{f,f'\in\calF} k_\calF(f,f')\leq \kappa$. We denote corresponding RKHS as $\calH_{\mathcal{F}}$ and canonical feature as $\phi_\mathcal{F}$. The inner product and the norm in $\calH_{\mathcal{F}}$ is denoted as $\braket[\calH_{\mathcal{F}}]{\cdot, \cdot}$  and $\|\cdot\|_{\calH_{\mathcal{F}}}$, respectively. We also define the tensor product as $\otimes$, which induces the Hilbert-Schmidt operator $f \otimes g:\calH_{\mathcal{F}_2}\to\calH_{\mathcal{F}_1}$ for $f\in\calH_{\mathcal{F}_1}, g\in\calH_{\mathcal{F}_2}$, such that $(f \otimes g)h = \braket[\calH_{\mathcal{F}_2}]{g,h}f$ for $h\in\calH_{\mathcal{F}_2}$. We denote the space of Hilbert-Schmidt operators of $\calH_{\mathcal{F}_2}\to\calH_{\mathcal{F}_1}$ as  $\calH_{\mathcal{F}_1\mathcal{F}_2}$ with Hilbert-Schmidt norm $\|\cdot\|_{\calH_{\mathcal{F}_1\mathcal{F}_2}}$. We introduce the notation of $\phi_{\calF_1\calF_2}(f_1, f_2) = \phi_{\calF_1}(f_1) \otimes \phi_{\calF_2}(f_2)$. Note that it can be shown that $\calH_{\mathcal{F}_1\mathcal{F}_2}$ is isometrically isomorphic to the product space $\calH_{\mathcal{F}_1} \times \calH_{\mathcal{F}_2}$.

\section{Identification} \label{sec:identification}
In this section, we show that the structural functions can be estimated from observable variables $(Y,A,W)$ given assumptions in \cref{sec:problem-settings}. Similar to classical PCL, we consider the bridge function whose partial average corresponds to the structural function.
\begin{thm}\label{thm:main-exitence}
    Given \cref{assum:structural,assum:completeness-confounder,assum:deterministic-confounding} and 
    \cref{assu:cond-exp-compactness,assu:cond-exp-L2,assu:cond-exp-compactness} presented in  \cref{sec:existence-of-bridge-function}, then there exists a bridge function $h_0(a,w)$ 
    such that 
    \begin{align}
    \expect{h_0(a, W)|A=a, Y=y} = y \quad \forall (a,y)\in\calA \times \calY\label{eq:bridge-function}.
    \end{align}
\end{thm}
Using the connection between our setting and PCL discussed in \cref{sec:connection-to-PCL}, \cref{thm:main-exitence} can be proved by generalizing PCL identification \citep{Miao2018Identifying}. The proof details can be found in  \cref{sec:identifiability-in-deterministic-confounding}. 

Under the sufficient condition discussed in \cref{thm:main}, we can obtain the structural function by taking the spatial average of bridge function
\begin{align}
    f_\struct(\tilde a) = \expect[W]{h_0(\tilde a,W)} \label{eq:partial-average-structrual}.
\end{align}
We first demonstrate this in the case where all variables follow linear structural equations and then move to a more general case. Finally, we conduct a theoretical analysis where the deterministic confounding assumption (\cref{assum:deterministic-confounding}) is violated.

\paragraph{Bridge function in the linear setting}
To illustrate our method and the implications of our assumptions, we first discuss the case where all the variables follow structural linear equations. Assume all variables are in $\mathbb{R}$ and follow: 
\begin{align*}    
    &U = \varepsilon_u, \, W = \alpha_{wu} U + \varepsilon_w, A = \alpha_{au} U + \varepsilon_a, \\
    &Y = \alpha_{ya} A + \alpha_{yu} U + \varepsilon_y,
\end{align*}
where $\varepsilon_u, \varepsilon_w, \varepsilon_a, \varepsilon_y$ are independent random variables with zero mean, and we assume all coefficients are non-zero. This setting is discussed by \citet{Kuroki2014Mesurement}, who show that we cannot recover $\alpha_{ya}$ unless we can estimate $\alpha^2_{wu}\mathrm{Var}[U]$. Our work avoids this requirement, however, by making the deterministic assumption. 
\begin{prop}\label{prop:linear-case}
    If we assume deterministic confounding $\varepsilon_y= 0$, the function
    \begin{align*}
        h_0(a,w) = \alpha_{ya} a +  (\alpha_{yu} / \alpha_{wu}) w
    \end{align*}
    satisfies  equation \eqref{eq:bridge-function}, and we have $\expect{h_0(\tilde a,W)} = f_\struct(\tilde a) = \alpha_{ya}\tilde a$.
\end{prop}
\begin{proof}
Let bridge function be $h_0(a,w) = h_a a +  h_w w$. Then, the left hand side of \eqref{eq:bridge-function} becomes 
\begin{align*}
    &\expect{h_0(a,W)|A=a, Y=y}\\
    &\quad = h_a a+ h_w\expect{W|A=a, Y=y}\\
    &\quad  = h_a a + h_w \alpha_{wu}  \expect{U|A=a, Y=y}\\
    &\quad  = h_a a + h_w \alpha_{wu}  (y - \alpha_{ya} a)/\alpha_{yu},
\end{align*}
where the last equation holds for deterministic confounding $\varepsilon_y = 0$. Hence, if we solve \eqref{eq:bridge-function}, we have $h_a = \alpha_{ya}, h_w = \alpha_{yu} / \alpha_{wu}$. Furthermore,  we have $\expect{h_0(a,W)} = \alpha_{ya}a$ since $\expect{W} = 0$.
\end{proof}

From \cref{prop:linear-case}, we can see that the structural function $f_\struct$ can be given as the partial average of bridge function $h_0(a, w)$ if we make the deterministic assumption. Note that \eqref{eq:bridge-function} only contains the observable variables, and thus we can estimate bridge function $h_0$ from samples $(A, Y, W)$ even in non-linear settings, as discussed in \cref{sec:method}.

\paragraph{Bridge function in the general case}

It is important to note that the solution of \eqref{eq:bridge-function} is not unique in general. Let $h_0$ be the ``true'' bridge function that satisfies \eqref{eq:partial-average-structrual}. Then all functions in 
\begin{align*}
    \mathcal{S} = \{h_0 + \delta(a,w)~|~ \expect{\delta(a,W)|A=a,Y=y} = 0\}
\end{align*}
satisfy \eqref{eq:bridge-function}, and some $h\in\mathcal{S}$ might give a different partial average $\expect{h(\tilde a,W)} \neq f_\struct(\tilde a)$. In the following theorem, we provide the sufficient condition that all partial averages in $\mathcal{S}$ correspond to the structural function.

\begin{thm}\label{thm:main}
For test point $\tilde a \in \calA$, given either $W \indepe A | Y^{(\tilde{a})}$ or 
    \begin{align}
    &\expect{l(U) \mid A=\tilde a, Y=y}=0\quad \forall y \in \mathcal{Y} \nonumber\\
    &\quad \Leftrightarrow l(u) = 0 \quad \prob{U}\text{-}\mathrm{a.e.}, \label{eq:informative-outcome}
\end{align}
then all functions in $\mathcal{S}$ satisfy $\expect{h(\tilde a, W)} = f_\struct(\tilde a)$.  
\end{thm}

The assumption $W \indepe A | Y^{(\tilde{a})}$ is employed by \citet{Tchetgen2023Single}, and ensures that the proxy $W$ only affects the treatment through the confounder $U$ whose mechanism can be captured by the potential output to be estimated $Y^{(\tilde a)}$. In \citep{Tchetgen2023Single}, it is discussed that municipality-level birth rates before the Zika virus outbreak satisfy the assumption when the treatment $A$ is the Zika epidemic status and the outcome $Y$ is the post-epidemic birth rate.
Our analysis extends the results of \citet{Tchetgen2023Single} to general treatments $A$, while only binary treatment $\calA =\{0,1\}$ is considered in the original analysis.

Another assumption in \eqref{eq:informative-outcome}  is similar to the ``informative treatment-proxy'' assumption in PCL; hence, we call it the ``informative outcome'' condition. This implies that the outcome can reveal the information on the hidden confounder.  The informative outcome assumption also holds in our chemical synthesis setting, since we can (in theory) recover the transient $U$ from the final $Y$ by solving the differential equation backward in time. Note that we do not need both $W \indepe A | Y^{(\tilde{a})}$ and \eqref{eq:informative-outcome}; either one is sufficient for identification.

The condition in \cref{thm:main} is not testable in practice and might be violated. However, our method appears robust to the violation, as shown in the empirical experiments in \cref{sec:experiment}.


\paragraph{Sensitivity Analysis on Deterministic Confounding} 
One natural question is whether we can recover the structural function when the outcome is given stochastically. In such a case, we cannot recover the bridge function $h_0$ by solving \eqref{eq:bridge-function}, and thus the estimate of the structural function is unavoidably biased.\footnote{We might be tempted to include the additive noise $\varepsilon$ in $Y = \gamma(A,U) + \varepsilon$ as part of the confounder $\tilde U = (U, \varepsilon)$, however this violates our modeling assumptions since the proxy $W$ is then not sufficiently informative for $\varepsilon$ (i.e. Assumption 2.2 fails for $\tilde U$). }
However, we can bound the bias if the noise is bounded and additive:

\begin{thm} \label{thm:main-stochastic}
    Assume that outcome is given as $Y = \gamma_0(A,U) + \varepsilon$ where $\expect{\varepsilon}=0,\, |\varepsilon|\leq M$. Suppose \cref{assum:structural} holds, and that there exists $\Xi \leq \infty$ for all bounded function $\ell(u)$,
    \begin{align}
    \sup_u |\ell(u)|\leq \Xi  \sup_{y\in\calY} {\expect{\ell(U)|A=a,Y=y}} \label{eq:dens-ratio}
    \end{align}
    for all $(a,y) \in \calA \times \calY$. If a
    bridge function $h$ satisfying \eqref{eq:bridge-function} exists, then for all $a\in\mathcal{A}$, we have  
    \begin{align*}
        |f_\struct(a) - \expect[W]{h(a,W)}| \leq M\Xi
    \end{align*}
\end{thm}

The proof can be found in \cref{sec:identifiability-in-stochastic-confounding}. Note that the condition in \eqref{eq:dens-ratio} is the sufficient condition for \eqref{eq:informative-outcome}. One important difference compared to \cref{thm:main} is that the existence of the bridge function $h_0(a,w)$ is not guaranteed in \cref{thm:main-stochastic}. This is because we need \cref{assum:deterministic-confounding} to prove the existence of the bridge function. Although \cref{thm:main-stochastic} only upper-bounds the error, the experiments in \cref{sec:experiment} show promising empirical results. Further theoretical analysis on this stochastic case is left as future work.

\section{Methods} \label{sec:method}
In this section, we present our proposed methods to estimate the bridge function. We introduce two algorithms, \emph{Single Kernel Proxy Variable (SKPV)} and \emph{Single Proxy Maximum Moment Restriction (SPMMR)}, which estimate the bridge function using different approaches.

From \cref{thm:main-exitence}, we can see that it is sufficient to estimate the bridge function to obtain the structural function, which can be done by solving functional equation \eqref{eq:bridge-function} using samples $\{a_i, w_i,y_i\}_{i=1}^n$. This problem is ill-posed when we consider a sufficiently rich functional space  \citep[see discussion of][]{Nashed1974Generalized}. Here, we assume that the bridge function $h$ is in RKHS $\calH_{\calA\calW}$ and add a smoothness requirement. We will discuss the case where the bridge function $h$ can be represented as neural network in \cref{sec:nn-methods}. 

We propose two approaches to solve \eqref{eq:bridge-function}: A \emph{Two-stage regression approach} \citep{Rahul2020KernelProxy,xu2021deep,Mastouri2021Proximal}, which minimizes the $L^2$-distance between both sides of \eqref{eq:bridge-function}; and a \emph{Maximum Moment Restriction approach} \citep{Mastouri2021Proximal,kompa2022deep,Kallus2021Causal}, which minimizes the maximum moment of the deviation in \eqref{eq:bridge-function}. We also derive novel closed-form solutions for these approaches, which improve the numerical stability compared to the existing solutions.

\subsection{Two-Stage Regression Approach}
In two-stage regression, we solve \eqref{eq:bridge-function} by minimizing the  loss
\begin{align*}
    &\twostageloss(h) = \expect[AY]{(Y\!-\!\expect{h(A,W)|A, Y})^2}\!+\!\eta \|h\|^2_{\calH_{\calA\calW}},
\end{align*}
where $\eta\|h\|^2_{\calH_{\calA\calW}}$ is the regularizing term. This loss cannot be directly minimized as it involves the conditional expectation $\expect{h(A,W)|A, Y}$. To deal with this,   \citetMastouri and \citet{Rahul2020KernelProxy}~employ a two-stage regression approach, which first estimates conditional expectation, and then minimizes the loss.
We apply this approach in our single proxy setting, and propose the \emph{Single Kernel Proxy Variable (SKPV)} method.  Specifically, if $h \in \calH_{\calA\calW}$, we have 
\begin{align*}
    &\expect{h(a,W)|A=a, Y=y}\\
    &\quad = \braket[\calH_{AW}]{h, \phi_\calA(a)\otimes \mu_{W|A,Y}(a,y)}.
\end{align*}
Here, $\mu_{W|A,Y}(a,y)$ is known as a \emph{conditional mean embedding} \citep{Song2009Hilbert,Gre2012CME,Li2022optimal,Park2020CME}, defined as $\mu_{W|A,Y}(a,y) = \expect{\phi_\calW(W)|A=a, Y=y}$. Under the regularity condition, 
there exists an operator $C_{W|A,Y}$ such that $\mu_{W|A,Y} = C_{W|A,Y}(\phi_\calA(a)\otimes\phi_\calY(y))$. This operator can be estimated from the samples $\{w_i, a_i, y_i\}_{i=1}^n$ \citep{Song2009Hilbert,Gre2012CME,Li2022optimal,Park2020CME} by minimizing the loss $\hat{C}_{W|A,Y} = \argmin_{C\in \calH_{\calW(\calA\calY)}} {\hat{\mathcal{L}}}^{\mathrm{cond}}(C)$ defined as 
\begin{align*}
    &{\hat{\mathcal{L}}}^{\mathrm{cond}}(C) \\
    &~=\frac1n \sum_{i=1}^n \|\phi_\calW(w_i) - C\phi_{\calA\calY}(a_i,y_i)\|_{\calH_\calW}^2 \!\!+\lambda\|C\|^2_{\calH_{\calW(\calA\calY)}}, 
\end{align*}
where $\phi_{\calA\calY}(a,y) = \phi_{\calA}(a) \otimes \phi_{\calY}(y)$ is the tensor feature, $\lambda>0$ is the regularizing coefficient, $\calH_{\calW(\calA\calY)}$ denotes the space of Hilbert-Schmidt operators that maps $\calH_{\calA\calY}$ to $\calH_{\calW}$, and $\|\cdot\|_{\calH_{\calW(\calA\calY)}}$ is the norm of that space. This is known as \emph{stage 1 regression} as we regress the feature $\phi_\calW(w_i)$ on tensor product features $\phi_\calA(a_i)\otimes\phi_\calY(y_i)$. The closed-from solution  is 
\begin{align*}
    \hat{C}_{W|A,Y} = \hat{C}_{W,(A,Y)}(\hat{C}_{(A,Y),(A,Y)} + \lambda I)^{-1}
\end{align*}
where $\hat{C}_{W,(A,Y)} = \frac1n \sum_{i=1}^n \phi_\calW(w_i)\otimes\phi_{\calA\calY}(a_i,y_i)$ and $\hat{C}_{(A,Y),(A,Y)} = \frac1n \sum_{i=1}^n \phi_{\calA\calY}(a_i,y_i)\otimes\phi_{\calA\calY}(a_i,y_i)$.
Given this, we minimize $\twostageloss$ using the estimate in stage 1 regression. Specifically, we obtain the bridge function estimate $\hat{h}$ as $\hat{h} = \argmin_{h\in\calH_{\calA\calW}} \hattwostageloss(h)$ given stage 2 data $\{\dot a_i, \dot y_i\}_{i=1}^m$, where the loss is defined as 
\begin{align*}
    &\hattwostageloss(h)=\\
    &~\frac1n \sum_{i=1}^n \left(\dot y_i-\braket[\calH_{\calA\calW}]{h, \phi_\calA(\dot a_i) \otimes \hat{C}_{W|A,Y}\phi_{\calA\calY}(\dot a_i,\dot y_i)}\right)^2\\
    &\quad +\eta\|h\|^2_{\calH_{\calA\calW}}.
\end{align*}
This regression is known as \emph{stage 2 regression}. We may use the same samples in stage 1 and 2 regressions, but it is reported that splitting samples between stage 1 and 2 regression can reduce bias in the estimation \citep{Joshua1995samplesplit}. Furthermore, since we do not need samples of $W$ in stage 2 regression, we may use an additional source of data for the samples $\{\dot a_i, \dot y_i\}$. We can obtain the closed-form solution of two-stage regression  as follows:

\begin{thm}\label{thm:two-stage-sol}
    Given stage 1 samples $\{w_i, a_i, y_i\}_{i=1}^n$ and stage 2 samples  $\{\dot{a}_i, \dot{y}_i\}_{i=1}^m$, and regularizing parameter $(\lambda, \eta)$, the minimizer of $\hattwostageloss$ is given as $\hat{h}(a,w) = \vec{\alpha}^\top \vec{k}(a,w)$ where
    \begin{align*}
        \vec{\alpha} = (M + m\eta I)^{-1}\dot{\vec{y}},~M = K_{\dot{A}\dot{A}} \odot (B^\top K_{WW} B).
    \end{align*}
    Here, $\dot{\vec{y}} = (\dot{y}_1,\dots, \dot{y}_m)^\top \in \mathbb{R}^m$, and $\vec{k}(a,w) = \vec{k}_{\dot{A}}(a) \odot (B^\top \vec{k}_{W}(w))\in \mathbb{R}^{m}$ where
    \begin{align*}
         B = (K_{AA}\odot K_{YY} + n\lambda I)^{-1}(K_{A{\dot{A}}}\odot K_{Y\dot{Y}}).
    \end{align*}
    For $F \in \{A,W,Y\}$, we denote the stage 1 kernel matrix as $K_{FF}= (k_\calF(f_i, f_j))_{ij} \in \mathbb{R}^{n\times n}$ where $k_\calF$ and $f_i$ are the corresponding space and stage 1 samples. Similarly, for $\dot{F} \in \{\dot{A},\dot{Y}\}$, we denote the stage 2 kernel matrix as $K_{\dot{F}\dot{F}}= (k_\calF(\dot{f}_i, \dot{f}_j))_{ij} \in \mathbb{R}^{n \times n}$ where $k_\calF$ and $f_i$ are the corresponding space and stage 2 samples. We further denote $K_{F\dot{F}} = (k_\calF(f_i, \dot{f}_j))_{ij} \in \mathbb{R}^{n\times m}$ and $\vec{k}_{\dot A}(a) = (k(\dot a_i, a))_i \in \mathbb{R}^m,  \vec{k}_{W}(w) = (k( w_i, w))_i \in \mathbb{R}^n$ 
\end{thm}

The proof is in \cref{sec:derivation}, where we further show that this solution is equivalent to the result of \citetMastouri[Proposition 2], which involves learning $nm$ parameters, while ours only requires learning $n$ parameters. Our solution is moreover slightly different from \citet[Algorithm 1]{Rahul2020KernelProxy}, which takes the form $\vec{\alpha} = (MM + m \eta M)^{-1}M\dot{\vec{y}}$. Although these solutions coincide when $M$ is invertible, we claim that our solution is more numerically stable, especially when  $M$ is close to singular, as shown in \cref{sec:stability-illustration}.  The computation requires $O(n^3 + m^3)$, which can be reduced via the usual Cholesky or Nystrom techniques. 

Given estimated bridge function $\hat{h}(\tilde a,w)$, we can estimate the structural function as 
$\hat{f}_\struct(\tilde a) = \frac1n \sum_{i=1}^n \hat{h}(\tilde a, w_i).$

\subsection{Maximum Moment Restriction Approach}
Another approach to solving \eqref{eq:bridge-function} is to consider moment restriction, which minimizes the maximum moment of $\prob{Y-h(a,W)|A,Y}$. We apply this approach to our setting and propose \emph{Single Proxy Maximum Moment Restriction (SPMMR)}. Specifically, we consider the following minimax problem.
\begin{align*}
    \argmin_{h\in\calH_{\calA\calW}}\max_{\substack{g\in\calH_{\calA\calY}, \|g\|_{\calH_{\calA\calY}}\leq 1}} (\expect{(Y-h(A,W))g(A,Y)})^2.
\end{align*}
As shown by \citetMastouri[Lemma 1], the unique solution to this minimax problem is the solution of \eqref{eq:bridge-function}. Furthermore, using a similar discussion to  \citetMastouri[Lemma 2], we can show that the maximum over $g$ can be computed in closed form as
\begin{align*}
    &\mmrloss(h)= \\
    &\quad \expect{\Delta_{Y,A,W}(h)\Delta_{Y',A',W'}(h)k_{\calA}(A,A')k_{\calY}(Y,Y')},
\end{align*}
where $\Delta_{Y,A,W}(h) = (Y-h(A,W))$ and $(W',A', Y')$ are independent copies of $(W, A, Y)$, respectively. Empirically, we minimize $\mmrloss$ with regularization as $\hat{h} = \argmin_{h\in\calH_{\calA\calW}} \hatmmrloss$, where
\begin{align*}
     \hatmmrloss &= \sum_{i,j=1}^n \frac{\Delta_i\Delta_j}{n^2}k_{\calA}(a_i,a_j)k_{\calY}(y_i,y_j)  \!+\! \eta \|h\|^2_{\calH_{\calA\calW}},
\end{align*}
and $\Delta_i = (y_i - h(a_i, w_i))$.
The closed-form solution of $\hat{h}$ can be obtained as follows:
\begin{thm}\label{thm:mmr-sol}
    Given samples $\{w_i, a_i, y_i\}_{i=1}^n$, the minimizer of $\hatmmrloss$ is given as $\hat{h}(a, w) = \vec{\alpha}^\top \vec{k}(a,w)$, where weight $\vec{\alpha}\in\mathbb{R}^n$ is  
    \begin{align*}
        \vec{\alpha} = \sqrt{G}\left(\sqrt{G}L\sqrt{G}+ n^2\eta I\right)^{-1}\sqrt{G}\vec{y},
    \end{align*}
    and
    \begin{align*}
         L = K_{AA} \odot K_{WW}, \quad G = K_{AA}\odot K_{YY}.
    \end{align*}
    Here, $\vec{y} = (y_1, \dots, y_n) \in \mathbb{R}^n$ and $\vec{k}(a,w) = (k_\calA(a_i,a)k_\calW(w_i,w))_{ij}\in\mathbb{R}^{n}$ with $K_{FF} = (k_\calF(f_i, f_j))_{ij} \in \mathbb{R}^{n\times n}$ for $F \in \{W,A,Y\}$ and corresponding kernel function $k_\calF$ and samples $f_i$, and $\sqrt{G}$ denotes the square root of $G= \sqrt{G}\sqrt{G}$.
\end{thm}
The derivation is given in \cref{sec:derivation}. Again, this slightly differs from the original solution in \citetMastouri, which is $\vec{\alpha} = (LGL + n^2 \eta L)^{-1}LG\vec{y}$, but they are identical if $L, G$ are non-singular. We claim that our solution is more numerically stable when the condition number of $L$ is large. The computational complexity is $O(n^3)$, which can be reduced via the usual Cholesky or Nystrom techniques. We can use the empirical average of the estimated bridge function to obtain the structural function as in SKPV. 

\subsection{Consistency}
Here, we state the consistency results for our estimates.

\begin{prop}\label{prop:2sr-consistency-simple}
    Assume \cref{assum:bridge-smooth,assum:operator-smooth} in \cref{sec:consistent-detail}. Given stage 1 samples $\{w_i, a_i, y_i\}_{i=1}^n$ and stage 2 samples  $\{\dot{a}_i, \dot{y}_i\}_{i=1}^m$,
    \begin{align*}
        \sup_{a,w} |h_0(a,w) - \hat{h}(a,w)| \to 0 
    \end{align*}
    with $n,m \to \infty$ by reducing regularizers $(\lambda, \eta)$ at appropriate rates.
\end{prop}
The proof is given in \cref{sec:consistent-detail}, in which we also provide the rate of convergence and the optimal values of regularizers $(\lambda, \eta)$. We can also show similar results for SPMMR.

\begin{prop}\label{prop:mmr-consistency-simple}
    Assume \cref{assum:mmr-smooth} in \cref{sec:consistent-detail}. Given samples $\{w_i, a_i, y_i\}$ size of $n$. Then,
    \begin{align*}
        \sup_{a,w} |\hat{h}(a,w) - h_0(a,w)|_{\calH_{\calA\calW}} \to 0
    \end{align*}
    as $n\to\infty$ by reducing the regularizer $\lambda$ at an appropriate rate.
\end{prop}
\cref{sec:consistent-detail} provides further details on this consistency result.

\section{Experiments} \label{sec:experiment}
In this section, we present the empirical performance of our single proxy method in a synthetic setting. Details of the experiments are summarized in \cref{sec:exp-detail} \footnote{Codes can be found at \url{https://github.com/liyuan9988/KernelSingleProxy/}}.

\begin{figure*}[t]
\begin{minipage}{0.68\textwidth}
    \centering
    \includegraphics[width=\textwidth]{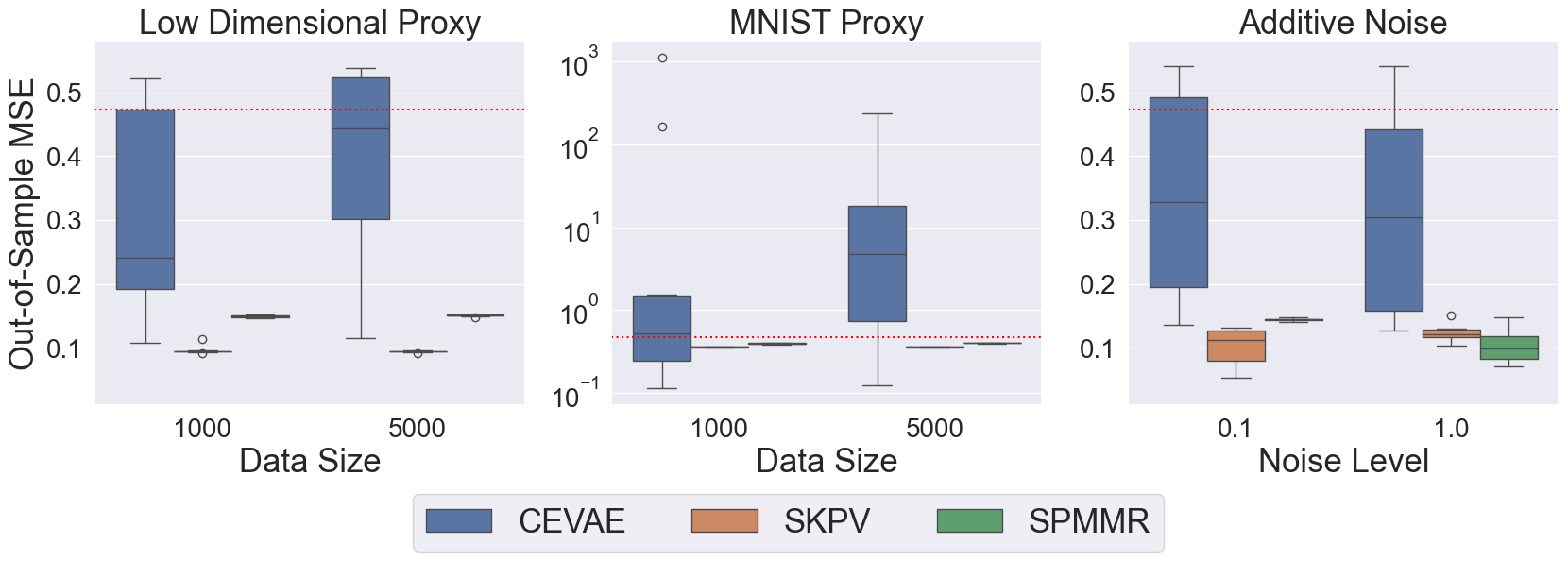}
    \caption{Results for experiments. The red dotted line shows the error  of the regression 
 $\|\expect{Y|A}- f_\struct(A)\|^2$.}
    \label{fig:experiment-result}
\end{minipage}
\hfill
    \begin{minipage}{0.28\textwidth}
     \centering
 \includegraphics[width=\textwidth
]{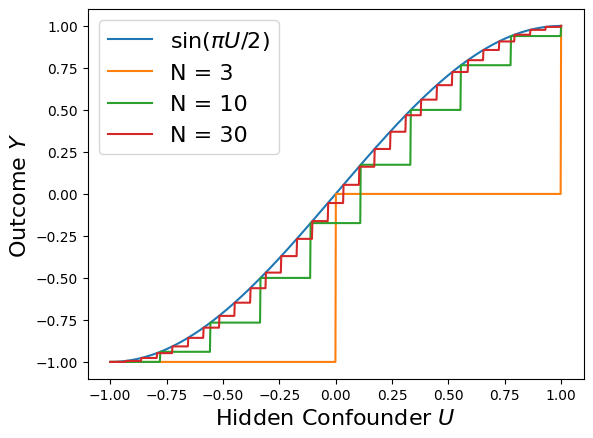}
 \caption{Plot of $\widetilde{\sin}_N(\pi U/2)$}
 \label{fig:disc_sine}
\end{minipage}
\end{figure*}

\paragraph{Low-Dimensional Proxy}
We consider the following data generation processes: We generate hidden confounder as $U \sim \mathrm{Unif}(-1, 1)$ and 
\begin{align}
     &A =\Phi(U) + \varepsilon_1, W = \exp(U) + \varepsilon_2, \nonumber\\
     & Y =\sin(\pi U/2) + A^2 - 0.3. \label{eq:org_data_gen}
\end{align}
Here, $\Phi$ is the Gaussian error function, and $\varepsilon_1 \sim \mathcal{N}(0, (0.1)^2), \varepsilon_2 \sim \mathcal{N}(0, (0.05)^2)$. The true structural function is $f_\struct(a) = a^2-0.3$. Note that deterministic confounding  \cref{assum:deterministic-confounding} and informative outcome assumption \eqref{eq:informative-outcome} hold in this setting since $U = \frac2{\pi} \arcsin(Y - A^2 -0.3)$.

To estimate the structural function, we apply SKPV and SPMMR. We used the Gaussian kernel, where the bandwidth is tuned using the median trick. We selected the regularizers using the procedure described in \cref{sec:exp-detail}. We split data evenly for stage 1 and 2 regression in SKPV. We compare proposed methods to CEVAE \citep{Louizos2017Causal}, which uses a VAE \citep{Kingma2014VAE} to recover the distribution of confounder $U$ from the ``proxy'' $W$. Although CEVAE does not always provide consistent causal effect estimates \citep{rissanen2021a}, it has shown strong empirical performance in a number of settings. The estimation loss is summarized in the first plot of \cref{fig:experiment-result}, in which we show the error of naive regression ignoring the confounding in the red line. In  \cref{fig:experiment-result}, SKPV and SPMMR consistently outperform CEVAE when the data size is set to 1000 and 5000. In our experiment, SKPV performs better than SPMMR. 

\paragraph{High-Dimensional Proxy}

We also consider the case where the proxy is high-dimensional. We used the same data generation process for $(A,Y,U)$  in \eqref{eq:org_data_gen} with deterministic confounding, and replaced the proxy $W$ with MNIST images, where the digit label is chosen as $\lfloor 5U + 5 \rfloor$. Results with data of size 1000 and 5000 are summarized in the second plot in \cref{fig:experiment-result}. Since the proxy $W$ is high-dimensional, all methods perform worse than in the low-dim proxy setting. In particular, the performance of CEVAE is highly unstable, consistent with the observation of \cite{rissanen2021a} that for high dimensional $W$, the autoencoder can prioritize quality of reconstruction of $W$ over causal effect estimation.
SKPV and SPMMR perform considerably better than CEVAE in this setting.

\paragraph{Sensitivity Analysis on Deterministic Assumption} We further test performance in a case where deterministic confounding  (\cref{assum:deterministic-confounding}) is violated. We added Gaussian noise $\mathcal{N}(0, \sigma^2)$ to the outcome $Y$ in \eqref{eq:org_data_gen}, where we varied the noise level in  $\sigma \in \{0.1, 1.0\}$. We again generated 1000 samples and repeated each noise level  10 times. The mean squared error in estimating the structural function is presented in the third plot in \cref{fig:experiment-result}. We can see that SKPV and SPMMR can still correct confounding bias even if the outcome contains noise. Furthermore, SPMMR performs better in the large noise case $\sigma=1.0$. We hypothesize that this is because the true structural function lies in support of data distribution once  noise is added, as shown in \cref{sec:exp-detail}. Overall, we see that the effect of the additive error is minimal, as suggested by \cref{thm:main-stochastic}.

\begin{figure}[h]
    \centering 
    \includegraphics[width=0.5\linewidth]{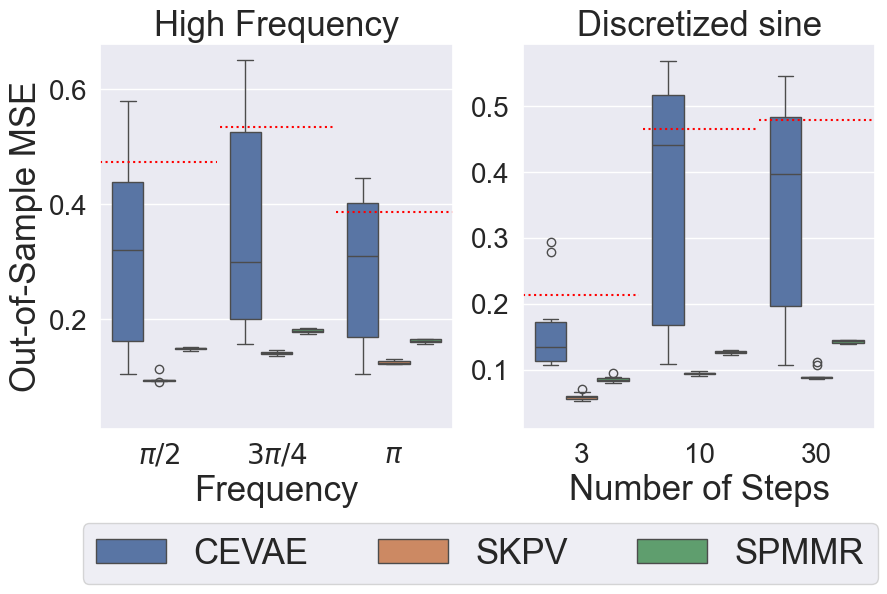}
    \caption{Result for sensitivity analysis on informative assumption.The red dotted line shows the error  of the regression 
 $\|\expect{Y|A}- f_\struct(A)\|^2$.}
    \label{fig:informative_assumption_res}
\end{figure}

\paragraph{Sensitivity Analysis on Informative Outcome Assumption}

To test sensitivity on the informative outcome assumption \eqref{eq:informative-outcome}, we
consider two settings. One is to increase the frequency of the dependence on $U$,
\begin{align*}
    Y = \sin(\beta U) + A^2 - 0.3,
\end{align*}
where we consider $\beta = \{\pi/2, 3\pi/4, \pi\}$. For $\beta > \pi/2$, the informative outcome assumption \eqref{eq:informative-outcome} is violated since there are multiple $U$ that can generate the same $Y$. The result is shown in the left plot of \cref{fig:informative_assumption_res}. This shows that there is indeed a marginal increase in error for both methods.

We next approximate the sine curve with the staircase function,
\begin{align*}
    Y = \widetilde{\sin}_N(\pi U / 2) + A^2 - 0.3,
\end{align*}
where $\widetilde{\sin}_N$ is the discretized sine curve as shown in \cref{fig:disc_sine}. Smaller $N$ means a wider range of $U$ corresponding to the same $Y$, hence the outcome $Y$ becomes less informative. We investigated each of  $N = \{3, 10, 30\}$ with 1000 samples;  results are shown in the right plot of \cref{fig:informative_assumption_res}, in which we can see the counter-intuitive results that smaller $N$ results in the better performance. This is because as we increase the number of steps, the confounding bias $\|\expect{Y|A} - f_\struct(A)\|^2$ decreases, as shown in the red dotted line in \cref{fig:informative_assumption_res}. This demonstrates a larger principle: the existence of settings where
the informative outcome assumption being ``more violated'' results in the outcome $Y$ being ``less dependent'' on the confounder $U$, which results in a smaller confounding bias. 
Note, however, that the situation is more complicated when frequency $\beta$ increases, with confounding bias first increasing, and then decreasing below that for the ``informative'' setting ($\beta=\pi/2$), as seen in \cref{fig:informative_assumption_res}.
In sum, a violation of the informativeness assumption \eqref{eq:informative-outcome} does not necessarily harm performance.


\section{Conclusion}
In this paper, we propose single proxy causal learning, which only requires a single proxy variable. Our main contribution is to extend \citet{Tchetgen2023Single} by drawing the connection with the original PCL, notably by identifying the outcome variable with the treatment proxy in the original PCL setting. This insight enables us to consider continuous treatments in the single proxy setting and use the alternative {\em informative outcome} assumption for the identification.

Following the theoretical work, we propose two methods,  SKPV and SPMMR, to estimate the bridge function which are shown to converge to the true bridge function as the number of samples increases. We also empirically show that we can still apply these methods in the additive noise case, and demonstrate a bound on the performance penalty incurred through violation of the deterministic outcome and informative outcome assumptions.

\section*{Acknowledgments}
This work was supported by the Gatsby Charitable Foundation.
We thank Ben Deaner for helpful discussions.

\bibliography{reference}

\onecolumn
\appendix

\section{Observable Confounder} \label{sec:observable-confounder}

\begin{figure}[h]
    \centering
    \begin{tikzpicture}
        \color{black}
        \node[state] (eps) at (0, 0) {$U$};
        \node[state, fill=yellow] (x) at (-1.8,-2) {$A$};
    
        \node[state, fill=yellow] (y) at (1.8, -2) {$Y$};
        \node[state, fill=yellow] (w) at (2.8, 0) {$W$};
        \node[state, fill=yellow] (obs) at (-2.8, 0) {$X$};
        \path[very thick] (x) edge (y);
        \path[very thick] (eps) edge (y);
        \path (eps) edge (x);
        \path (obs) edge (x);
        \path[very thick] (obs) edge (y);
        \path[bidirected] (eps) edge (w);
        \path[bidirected] (obs) edge (eps);
    \end{tikzpicture} 
    \caption{Typical causal graph for the single proxy variable with observable confounder $X$. Here, the bidirectional arrows mean that we allow an arrow in either direction, or even a common ancestor variable; and the thick arrow indicates a deterministic relationship between the variables. }
    \label{fig:causal-graph-obs}
\end{figure}

In this appendix, we discuss the case where there exists the observable confounder $X$. The causal graph in this case is given in \cref{fig:causal-graph-obs}. Now, the structural function is defined as 
\begin{align*}
    f_\struct(\tilde{a}) = \expect[X,U]{\expect{Y|A=\tilde a, X, U}}.
\end{align*}
Given the $X \in \mathcal{X}$, we would solve the bridge function $h_0(a,x,w): \mathcal{A}\times\mathcal{X}\times\mathcal{W}\to \mathcal{Y}$,
\begin{align}
    \expect{h_0(a,x,W)|A=a, X=x, Y=y} = y \quad \forall a \in \mathcal{A}, y \in \mathcal{Y}, x\in\mathcal{X} \label{eq:bridge-function-obs}
\end{align}
Such bridge function $h_0$ exists when the proxy satisfies similar conditions.
\begin{assum} \label{assum:structural-obs}
We assume $W \indepe (A,Y) | U, X$.
\end{assum}
\begin{assum} \label{assum:completeness-confounder-obs}
For any square integrable function $l: \mathcal{U} \to \mathbb{R}$, the following conditions hold for all $a \in \mathcal{A}$ and $x\in\mathcal{X}$:
\begin{align*}
    &\expect{l(U) \mid A= a, W=w, X=x}=0\quad \forall w \in \mathcal{W}\quad  \Leftrightarrow \quad l(u) = 0  \quad \prob{U}\text{-}\mathrm{a.e.}
\end{align*}
\end{assum}

\begin{assum} \label{assum:deterministic-confounding-obs}
    There exists a \emph{deterministic} function $\gamma_0(a,x, u)$ such that $Y = \gamma_0(A,U,X)$ and $\|\gamma_0(a,x, U)\|_{\prob{U}} \leq \infty$ for all $a\in\calA, x \in \mathcal{X}$.
\end{assum}
We can show the existence using a similar discussion to \cref{thm:main-exitence}. The structural function is obtained by the partial average over the bridge function as $\expect{h(\tilde a,X,W)} = f_\struct(\tilde a)$ if we further assume either of the following:
\begin{assum}[
 {\citealp[Assumption 2(\romannumeral 3)]{Tchetgen2023Single}}] \label{assum:coca-assum-obs}
    For test point $\tilde a \in \mathcal{A}$, we have $W \indepe A | X, Y^{(\tilde{a})}$.
\end{assum}
\begin{assum} \label{assum:informative-outcome-obs}
For any square integrable function $l:\calU\to\mathbb{R}$, the following conditions hold for test point $\tilde a\in\mathcal{A}$ and for any $x\in\mathcal{X}$:
\begin{align*}
    &\expect{l(U) \mid A=\tilde a, Y=y, X=x}=0\quad \forall y \in \mathcal{Y} \quad\Leftrightarrow \quad l(u) = 0 \quad \prob{U}\text{-}\mathrm{a.e.}
\end{align*}
\end{assum}

We may use either two-stage regression or the maximum moment restriction approach to obtain the bridge function. We state the closed-form solution for each approach.
\begin{thm}\label{thm:two-stage-sol-obs}
    Given stage 1 samples $\{w_i, a_i, y_i, x_i\}_{i=1}^n$ and stage 2 samples  $\{\dot{a}_i, \dot{y}_i, \dot{x}_i\}_{i=1}^m$, and regularizing parameter $(\lambda, \eta)$, the solution of two-stage regression approach is given as 
    \begin{align*}
        \hat{h}(a,w,x) = \vec{\alpha}^\top \vec{k}(a,w,x),\quad \vec{\alpha} = (M + m\eta I)^{-1}\dot{\vec{y}}
    \end{align*}
    where $\dot{\vec{y}} = (\dot{y}_1,\dots, \dot{y}_m)^\top \in \mathbb{R}^m$, and $M \in \mathbb{R}^{m\times m}$ and $\vec{k}(a,w) \in \mathbb{R}^{m}$ are defined as 
    \begin{align*}
        &M = K_{\dot{A}\dot{A}} \odot (B^\top K_{WW} B) \odot K_{\dot{X}\dot{X}},\\
        &\vec{k}(a,w, x) = \vec{k}_{\dot{A}}(a) \odot (B^\top \vec{k}_{W}(w)) \odot \vec{k}_{\dot{X}}(x)\\
        &B = (K_{AA}\odot K_{YY} \odot K_{XX}+ n\lambda I)
        ^{-1}(K_{A{\dot{A}}}\odot K_{Y\dot{Y}} \odot K_{X\dot{X}}).
    \end{align*}
    Here, for $F \in \{A,W,Y,X\}$, we denote the stage 1 kernel matrix as $K_{FF}= (k_\calF(f_i, f_j))_{ij} \in \mathbb{R}^{n\times n}$ where $k_\calF$ and $f_i$ are the corresponding space and stage 1 samples. Similarly, for $\dot{F} \in \{\dot{A},\dot{Y},\dot{X}\}$, we denote the stage 2 kernel matrix as $K_{\dot{F}\dot{F}}= (k_\calF(\dot{f}_i, \dot{f}_j))_{ij} \in \mathbb{R}^{n \times n}$ where $k_\calF$ and $f_i$ are the corresponding space and stage 2 samples. We further denote $K_{F\dot{F}} = (k_\calF(f_i, \dot{f}_j))_{ij} \in \mathbb{R}^{n\times m}$ and $\vec{k}_{\dot A}(a) = (k(\dot a_i, a))_i \in \mathbb{R}^m,  \vec{k}_{W}(w) = (k( w_i, w))_i \in \mathbb{R}^n, \vec{k}_{\dot X}(s) = (k(\dot x_i, x))_i \in \mathbb{R}^m$ 
\end{thm}

\begin{thm}\label{thm:mmr-sol-obs}
    Given samples $\{w_i, a_i, y_i, x_i\}_{i=1}^n$, the solution of maximum moment restriction approach is given as $\hat{h}(a, w) = \vec{\alpha}^\top \vec{k}(a,w)$, where weight $\vec{\alpha}\in\mathbb{R}^n$ is given as 
    \begin{align*}
        \vec{\alpha} = \sqrt{G}\left(\sqrt{G}L\sqrt{G}+ n^2\eta I\right)^{-1}\sqrt{G}\vec{y}.
    \end{align*}
    Here, $\vec{y} = (y_1, \dots, y_n) \in \mathbb{R}^n$ and 
    \begin{align*}
        &L = K_{AA} \odot K_{WW} \odot K_{XX}, \quad G = K_{AA}\odot K_{YY} \odot K_{XX},\\
        &\vec{k}(a,w) = (k_\calA(a_i,a)k_\calW(w_i,w)k_{\mathcal{X}}(x_i,x))_{i}\in\mathbb{R}^{n}
    \end{align*}
    with $K_{FF} = (k_\calF(f_i, f_j))_{ij} \in \mathbb{R}^{n\times n}$ for $F \in \{W,A,Y,X\}$ and corresponding kernel function $k_\calF$ and samples $f_i$, and $\sqrt{G}$ denotes the square root of $G= \sqrt{G}\sqrt{G}$.
\end{thm}

The consistency of each estimation can be proved by a similar discussion presented in \cref{prop:2sr-consistency} and \cref{prop:mmr-consistency}.
\section{Connection to PCL} \label{sec:connection-to-PCL}
In Proxy Causal Learning (PCL), we assume access to a treatment-inducing proxy variable $Z$, and an outcome-inducing proxy variable $W$, which satisfy the following \textit{structural assumption} and \textit{completeness assumption}.
\begin{assum}[Structural Assumption \citep{Deaner2018Proxy,Mastouri2021Proximal}] \label{assum:stuctural-pcl}
We assume $Y \indepe Z | A, U$, and $W \indepe (A,Z) | U$.
\end{assum}
\begin{assum}[Completeness Assumption on Confounder \citep{Deaner2018Proxy,Mastouri2021Proximal}] \label{assum:completeness-confounder-pcl}
Let $l: \mathcal{U} \to \mathbb{R}$ be any square integrable function $\|l\|_{\prob{U}} < \infty$.
The following conditions hold for all $a \in \mathcal{A}$
\begin{align*}
    &\expect{l(U) \mid A=a, W=w}=0\quad \forall w \in \mathcal{W} \quad\Leftrightarrow \quad l(u) = 0 \quad \prob{U}\text{-}\mathrm{a.e.}\\
    &\expect{l(U) \mid A=a, Z=z}=0\quad \forall z \in \mathcal{Z} \quad\Leftrightarrow \quad l(u) = 0 \quad \prob{U}\text{-}\mathrm{a.e.}
\end{align*}
\end{assum}
Here, $A,Y,U$ are the treatment, the outcome, and the unobserved confounder, respectively. We show that given \cref{assum:bridge-smooth,assum:completeness-confounder,assum:deterministic-confounding} and informative outcome assumption \eqref{eq:informative-outcome}, these PCL assumptions hold with $Z = Y$.

Since $Y$ is deterministic given $A,U$ (\cref{assum:deterministic-confounding}), we have
\begin{align*}
    Y \indepe Y | A, U
\end{align*}
since the deterministic variables are independent of the deterministic variables by definition. Also, from \cref{assum:structural}, we have $W \indepe (Z,Y) |U$. Therefore, we have \cref{assum:stuctural-pcl}. Furthermore, the first line of \cref{assum:completeness-confounder-pcl} is equivalent to \cref{assum:completeness-confounder}, and the second line of \cref{assum:completeness-confounder-pcl} corresponds to informative outcome assumption \eqref{eq:informative-outcome} with $Z = Y$.

From this, we show that the outcome $Y$ can be used as the proxy $Z$ in PCL. We use this connection to derive the functional equation for bridge function \eqref{eq:bridge-function} and its relation to structural function $f_\struct$. In PCL, the bridge function $h_0$ is defined as the solution of 
\begin{align*}
    \expect{h_0(a,W)|A, Z} = \expect{Y|A, Z},
\end{align*}
and we have $f_\struct(a) = \expect[W]{h_0(a,W)}$.
Now, if we replace the proxy $Z$ to the outcome $Y$, the right-hand side becomes just $Y$, and the left-hand side is $\expect{h_0(a,W)|A, Y}$, which gives the definition in our bridge function \eqref{eq:bridge-function}. From this, we can derive the identification results (\cref{thm:main}) by using a similar discussion to the PCL identification.
\section{Numerical Stability} \label{sec:stability-illustration}
\begin{figure}
    \centering
    \includegraphics[width=0.5\linewidth]{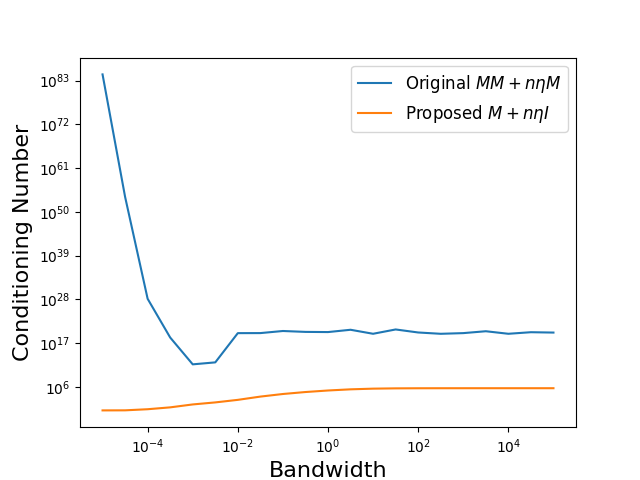}
    \caption{Conditioning number for different SKPV formulations. We tested on our deterministic low-dimensional proxy setting with 1000 samples.}
    \label{fig:conditioning-number}
\end{figure}

To illustrate the numerical stability of our solution, we plot the conditioning numbers with respect to the various bandwidths of the Gaussian kernel in our synthetic experimental setting in \cref{fig:conditioning-number}. This shows matrix to be inverted in our solution has a smaller conditioning number (orange) than that in the original solution (blue). We observed similar phenomena in SPMMR.

\section{Use of Neural Network Features} \label{sec:nn-methods}
A number of works \citep{xu2021deep,kompa2022deep,Kallus2021Causal} proposed the deep learning methods for proxy causal learning, which can be applied to our settings by using the connection discussed in \cref{sec:connection-to-PCL}. We expect these methods to achieve better empirical performances than kernel-based methods, especially when the data is high-dimensional as discussed in the original PCL setting \citep{xu2021deep,kompa2022deep,Kallus2021Causal}. We first discuss the two-stage regression approach using neural nets, and then introduce the neural version of the maximum moment restriction approach.

\paragraph{Two-stage Regression Approach}
We can apply a similar method as DFPV \citep{xu2021deep} to derive the neural version of the two-stage regression approach. In DFPV, we model the bridge function $h_0(a,w)$ as 
\begin{align*}
    h(a,w) = \vec{u}^\top (\vec{\psi}_{\theta_{A(2)}}(a) \otimes \vec{\psi}_{\theta_{W}}(w)), 
\end{align*}
where $\vec{u}$ is weight parameters, $\otimes$ is the tensor product $\vec{a}\otimes\vec{b} = \mathrm{vec}(ab^\top)$ and $\vec{\psi}_{\theta_{A(2)}}, \vec{\psi}_{\theta_{W}}$ are neural nets parametrized by $\theta_{A(2)}, \theta_{W}$, respectively. We minimize the loss $\twostageloss$ to learn these parameters.
\begin{align*}
    \twostageloss(h) = \expect[AY]{(Y - \vec{u}^\top (\vec{\psi}_{\theta_{A(2)}}(a) \otimes \expect{\vec{\psi}_{\theta_{W}}(W)|A,Y}))}
\end{align*}
Similar to the SKPV, we need to model the conditional feature mean $\expect{\vec{\psi}_{\theta_{W}}(W)|A,Y}$ to minimize the loss $\twostageloss$. We model this using the different neural net feature maps.
\begin{align*}
\expect[W|A=a,Y=y]{\vec{\psi}_{\theta_{W}}(W)} = \vec{V}(\vec{\phi}_{\theta_{A(1)}}(a) \otimes \vec{\phi}_{\theta_Y}(y)), 
\end{align*}
where $\vec{V}$ are parameters, and $\vec{\phi}_{\theta_{A(1)}}, \vec{\phi}_{\theta_Z}$ are neural nets parametrized by $\theta_{A(1)}, \theta_Z$, respectively. Note we may use different neural nets in the treatment features $\vec{\phi}_{\theta_{A(1)}}$ and $\vec{\psi}_{\theta_{A(2)}}$.

As in SKPV, we learn $\expect[W|a,y]{\vec{\psi}_{\theta_{W}}(W)}$ in stage 1 regression and $h_0(a,w)$ in stage 2 regression, but in addition to the weights $\vec{u}$ and $\vec{V}$, we also learn the parameters of the feature maps. 

Specifically, in stage 1, we learn $\vec{V}$ and parameters $\theta_{A(1)}, \theta_Z$ by minimizing the following empirical loss:
\begin{align*}
     {\hat{\mathcal{L}}}^{\mathrm{cond}}(\vec{V}, \theta_{A(1)}, \theta_Z) = \frac1m \sum_{i=1}^m \left\|\vec{\psi}_{\theta_{W}}(w_i) - \vec{V}\left(\vec{\phi}_{\theta_{A(1)}}(a_i) \otimes\vec{\phi}_{\theta_Y}(y_i)\right) \right\|^2 + \lambda\|\vec{V}\|^2.
\end{align*}

Although ${\hat{\mathcal{L}}}^{\mathrm{cond}}$ depends on $\theta_{W}$, at this stage, we do not optimize $\theta_{W}$ with ${\hat{\mathcal{L}}}^{\mathrm{cond}}$ as ${\vec{\psi}_{\theta_{W}}(w)}$ is the  ``target variable'' in stage 1 regression. Given the minimizers $(\hat{\vec{V}}, \hat{\theta}_{A(1)}, \hat{\theta}_{Z}) = \argmin \hat{\mathcal{L}}_1$, we learn weights $\vec{u}$ and parameters $\theta_W, \theta_{A(2)}$ by minimizing the empirical stage 2 loss,
\begin{align*}
     {\hat{\mathcal{L}}}^{\mathrm{2SR}}(\vec{u}, \theta_W, \theta_{A(2)}) = \frac1n \sum_{i=1}^n \left(\dot y_i - \vec{u}^\top \left(\vec{\psi}_{\theta_{A(2)}}(\dot a_i) \otimes \hat{\vec{V}}  \left(\vec{\phi}_{\hat{\theta}_{A(1)}}(\dot a_i) \otimes\vec{\phi}_{\hat{\theta}_Y}(\dot y_i)\right)\right)\right)^2 + \eta\|\vec{u}\|^2.
\end{align*}
Although the expression of ${\hat{\mathcal{L}}}^{\mathrm{2SR}}$ does not explicitly contain $\theta_W$, 
it implicitly depends on $\theta_W$ as $(\hat{\vec{V}}, \hat{\theta}_{A(1)}, \hat{\theta}_{Z})$ is the solution of a minimization problem involving $\theta_W$. To deal with this implicit dependency, we may use the method proposed in \citet{xu2021deep}, in which we ignore the dependency of $\theta_W$ on parameters $\hat{\theta}_{A(1)}, \hat{\theta}_{Y}$, and compute the gradient via the closed-form solution of $\hat{V}$.

This gives the following learning procedure. First, we fix parameters in the adaptive feature maps $(\theta_{A(1)}, \theta_Y, \theta_{A(2)}, \theta_{
W})$. Given these parameters, optimal weights $\hat{\vec{V}}, \hat{\vec{u}}$ can be learned by minimizing the empirical stage 1 loss ${\hat{\mathcal{L}}}^{\mathrm{cond}}$  and empirical stage 2 loss ${\hat{\mathcal{L}}}^{\mathrm{2SR}}$, respectively. These minimizations can be solved analytically, where the solutions are
\begin{align}
    &\hat{\vec{V}}(\vec{\theta}) = \Psi_1^\top \Phi_1 (\Phi_1^\top \Phi_1 + m\lambda_1 I)^{-1}, &
    &\hat{\vec{u}}(\vec{\theta}) = \left(\Phi_2 ^\top \Phi_2 + n\lambda_2 I\right)^{-1}\Phi_2^\top \vec{y}_2, \label{eq:weights-sol}
\end{align}
where we denote $\vec{\theta} = (\theta_{A(1)}, \theta_Y, \theta_{A(2)}, \theta_{W})$ and define matrices as follows:
\begin{align*}
    &\Psi_1 = \left[\vec{\psi}_{\theta_W}(w_1), \dots, \vec{\psi}_{\theta_W}(w_m)\right]^\top, & &\Phi_1 = [\vec{v}_1(a_1, y_1), \dots, \vec{v}_1(a_m, y_m)]^\top,\\
    &\vec{y}_2 = [\dot y_1, \dots, \dot y_n]^\top, & &\Phi_2 = [\vec{v}_2(\dot a_1, \dot y_1), \dots, \vec{v}_2(\dot a_n, \dot y_n)]^\top,&\\
    &\vec{v}_1(a, y) = \vec{\phi}_{\theta_{A(1)}}(a) \otimes\vec{\phi}_{\theta_Y}(y), &&\vec{v}_2(a, y) = \vec{\psi}_{\theta_{A(2)}}(a) \otimes \left(\hat{\vec{V}}(\vec{\theta}) \left(\vec{\phi}_{\theta_{A(1)}}(a) \otimes\vec{\phi}_{\theta_Y}(y)\right)\right).
\end{align*}
Given these weights $\hat{\vec{u}}(\vec{\theta}), \hat{\vec{V}}(\vec{\theta})$, we can update feature parameters by a gradient descent method with respect to the residuals of the loss of each stage, while regrading $\hat{\vec{V}}$ and $\hat{\vec{u}}$ as functions of parameters. Specifically, we take the gradient of the losses
\begin{align*}
    &{\hat{\mathcal{L}}}^{\mathrm{cond}}(\vec{\theta}) = \frac1m \sum_{i=1}^m \left\|\vec{\psi}_{\theta_{W}}(w_i) - \hat{\vec{V}}(\vec{\theta})\left(\vec{\phi}_{\theta_{A(1)}}(a_i) \otimes\vec{\phi}_{\theta_Y}(y_i)\right) \right\|^2 + \lambda \|\hat{\vec{V}}(\vec{\theta})\|^2,\\
    &{\hat{\mathcal{L}}}^{\mathrm{2SR}}(\vec{\theta}) = \frac1n \sum_{i=1}^n \left(\dot y_i - \hat{\vec{u}}(\vec{\theta})^\top \left(\vec{\psi}_{\theta_{A(2)}}(\dot a_i) \otimes \hat{\vec{V}}(\vec{\theta})  \left(\vec{\phi}_{\theta_{A(1)}}(\dot a_i) \otimes\vec{\phi}_{\theta_Y}(\dot y_i)\right)\right)\right)^2 \!+\!  \eta \|\hat{\vec{u}}(\vec{\theta})\|^2,
\end{align*}
where $\hat{\vec{V}}(\vec{\theta}), \hat{\vec{u}}(\vec{\theta})$ are given in \eqref{eq:weights-sol}. Given these losses, $(\theta_{A(1)}, \theta_{Y})$ are minimized with respect to ${\hat{\mathcal{L}}}^{\mathrm{cond}}$, and $(\theta_{A(2)}, \theta_{W})$ are minimized with respect to ${\hat{\mathcal{L}}}^{\mathrm{2SR}}$. \citet{xu2021deep} reported that the learning procedure is stabilized by running several gradient descent steps on the stage 1 parameters $(\theta_{A(1)}, \theta_{Y})$ before updating the stage 2 features $(\theta_{A(2)}, \theta_{W})$.

\paragraph{Maximum Moment Restriction}
In maximum moment restriction, we minimize the following loss to obtain the bridge function
\begin{align*}
    \hat{h} = \argmin_{h}\max_{g}(\expect{(Y-h(A,W))g(A,Y)})^2.
\end{align*}
In SPMMR, we assume both $h$ and $g$ to be in RKHS. There are two approaches to extend this to the neural network model. One is to model $h$ as a neural network while still assuming $g$ is in RKHS, which is considered in \citet{kompa2022deep}. The other is to model both $h,g$ as neural network functions as in \citet{Kallus2021Causal}.

If we assume $g$ to be the RKHS function, we can learn the bridge function by minimizing the same loss $\mmrloss$ to SPMMR.
\begin{align}
    \hat{h} = \argmin_h \expect{\Delta_{Y,A,W}(h)\Delta_{Y',A',W'}(h)k_{\calA}(A,A')k_{\calY}(Y,Y')}, \label{eq:mmr-loss-nn}
\end{align}
where $\Delta_{Y,A,W}(h) = (Y-h(A,W))$ and $(W',A', Y')$ are independent copies of $(W, A, Y)$, respectively. Empirically, we may add regularizer $\Omega$ and minimize the following loss.
\begin{align*}
    \hat{h} = \argmin_h \frac1{n^2} \sum_{i,j=1}^n \Delta_i(h)\Delta_j(h)k_{\calA}(a_i,a_j)k_{\calY}(y_i,y_j) + \eta \Omega(h)
\end{align*}
\citet{kompa2022deep} suggested to use the $L^2$-penalty on the parameter in $h$ as the regularizer.

When we model $g$ as a neural net function, we may still obtain the bridge function $h$ by solving \eqref{eq:mmr-loss-nn}. However, \citet{Kallus2021Causal} proposed the following loss, which is more computationally efficient.
\begin{align*}
    \hat{h} = \argmin_h \max_g \frac1n \sum_{i=1}^n (y_i - h(a_i,w_i))g(a_i, y_i) - \frac{\lambda}n \sum_{i=1}^n g^2(a_i, y_i)
\end{align*}
Note that the objective of minimax problem can be computed in $O(n)$ time, while the original loss \eqref{eq:mmr-loss-nn} requires $O(n^2)$.
The second term is called stabilizer \citep{Kallus2021Causal}. \citet{Kallus2021Causal} showed that stabilizers are different from regularizers, and generally should not let it vanish when the sample size grows. 

\section{Details of Identifiability}\label{sec:identifiability-detail}
In this appendix, we prove the propositions given in the main text.

\subsection{Existence of bridge function} \label{sec:existence-of-bridge-function}
Here, we present the proof of \cref{thm:main-exitence}. We introduce the following operator
Let us consider the following operators:
\begin{align*}
E_{a}:L^{2}(P_{W\given A=a})\to L^{2}(P_{Y|A=a}),\ E_{a}f & \coloneqq\expect{f(W)\given A=a,Y=\cdot},\\
F_{a}:L^{2}(P_{Y\given A=a})\to L^{2}(P_{W|A=a}),\ F_{a}g & \coloneqq\expect{g(Y)\given A=a,W=\cdot},
\end{align*}
Our goal is to show that $I(y)=y$ is in the range of
$E_{a},$ i.e., we seek a solution to the inverse problem defined
by
\begin{equation}
E_{a} h= I(y).\label{eq:inteq}
\end{equation}

This suffices to prove the existence of the function $h_a$ for if
there exists a function $h^*_{a}$ for each $a\in{\cal A}$ such that
\[
\expect{h^*_a|A=a,Y=\cdot}=y,
\]
we can define $h_0(a,w)\coloneqq h^*_{a}(w)$. We apply the discussion similar to \citet[Appendix B]{xu2021deep}, which ensures the existence using the following theorem.

\begin{prop}[{\citealp[Theorem 15.18]{Kress1999linear}}] \label{prop:picard}Let $\mathcal{X}$ and $\mathcal{Y}$ be Hilbert spaces.
Let $E:\mathcal{X}\to \mathcal{Y}$ be a compact linear operator with singular system $\{(\mu_{n},\varphi_{n},g_{n})\}_{n=1}^{\infty}.$
The equation of the first kind 
\[
E\varphi=f
\]
 is solvable if and only if $f\in N(E^{*})^{\perp}$ and 
\[
\sum_{n=1}^{\infty}\frac{1}{\mu_{n}^{2}}\lvert\la f,g_{n}\ra\rvert^{2}<\infty.
\]
Here, $N(E^*)$ denotes the null space of the operator $E^*$. 
 Then a solution is given by 
\[
\phi=\sum_{n=1}^{\infty}\frac{1}{\mu_{n}}\la f,g_{n}\ra\varphi_{n}.
\]
\end{prop}
To apply Proposition \ref{prop:picard}, we make the following additional
assumptions.
\begin{assum}
\label{assu:cond-exp-compactness} For each $a\in{\cal A},$ the operator
$E_{a}$ is compact with singular system $\{(\mu_{a,n},\varphi_{a,n},g_{a,n})\}_{n=1}^{\infty}.$ 
\end{assum}

\begin{assum}
\label{assu:cond-exp-L2} The identity map $I(y)$ satisfies
\[
\sum_{n=1}^{\infty}\frac{1}{\mu_{a,n}^{2}}\lvert\la I ,g_{a,n}\ra_{L^{2}(P_{Y\given A=a})}\rvert^{2}<\infty,
\]
for a singular system $\{(\mu_{a,n},\phi_{a,n},g_{a,n})\}_{n=1}^{\infty}$
given in Assumption \ref{assu:cond-exp-compactness}. 
\end{assum}

Please refer to \citet[Remark 1]{xu2021deep} for the discussion on these assumptions. It is easy to see that \cref{assu:cond-exp-compactness,assu:cond-exp-L2} are required for using Proposition~\ref{prop:picard}. The remaining condition
to show is that $I(y)$ is in $N(E_{a}^{*})^{\perp}.$
We show that the structural assumption (Assumption~\ref{assum:structural}), completeness assumption (Assumption~\ref{assum:completeness-confounder}), and deterministic confounding (\cref{assum:deterministic-confounding}) imply the required condition.

\begin{lem}
\label{lem:null-space-assumption} Under \cref{assum:structural,assum:completeness-confounder,assum:deterministic-confounding},
the identity $I(y)$ is in the
orthogonal complement of the null space $N(E_{a}^{*}).$
\begin{proof}
We first show that the adjoint of $E_{a}$ is given by $F_{a}.$ For
the operator $E_{a},$ any $f\in L_{2}(P_{W|A=a})$ and $g\in L_{2}(P_{Y|A=a}),$
we have 
\begin{align*}
\inner{E_{a}f}g_{L_{2}(P_{Y\given A=a})} & =\expect[Y|A=a]{\expect{f(W)\given A=a,Y}g(Y)}\\
& =\expect[Y|A=a]{\expect[U|A=a, Y]{\expect{f(W)\given A=a,Y,U}}g(Y)}\\
& \overset{\mathrm{(a)}}{=}\expect[Y|A=a]{\expect[U|A=a, Y]{\expect{f(W)\given A=a,U}}g(Y)}\\
& =\expect[U, Y|A=a]{\expect{f(W)\given A=a,U}g(Y)}\\
& =\expect[U|A=a]{\expect{f(W)\given A=a,U}\expect{g(Y)\given A=a,U}}
\end{align*}
where (a) follows from $W \indepe (A,Y) | U$, which is from Assumption~\ref{assum:structural}. Similarly,
\begin{align*}
 & \inner f{F_{a}g}_{L^{2}(P_{W|A=a})}=\expect[W|A=a]{f(W) \expect{g(Y)\given A=a, W}}\\
 &=\expect[W|A=a]{f(W) \expect[U|A=a, W]{\expect{g(Y)\given A=a, W, U}}}\\
 &\overset{\mathrm{(b)}}{=} \expect[W|A=a]{f(W) \expect[U|A=a, W]{\expect{g(Y)\given A=a, U}}}\\
 &=\expect[W,U|A=a]{f(W)\expect{g(Y)\given A=a, U}}\\
 &=\expect[U|A=a]{\expect{f(W)\given A=a, U} \expect{g(Y)\given A=a, U}}\\
 & =\inner{E_{a}f}g_{L_{2}(P_{Y\given A=a})}.
\end{align*}
Again, (b) is given by $W \indepe A,Y | U$ from Assumption~\ref{assum:structural}.
For any $f^{*}\in N(E^*_{a}) = N(F_a),$ by iterated expectations, we have 
\begin{align}
0 & =\expect{f^{*}(Y)|A=a,W=\cdot}\nonumber \\
 & =\expect[U]{\expect{f^{*}(Y)\given A,U,W}\given A=a,W=\cdot}\nonumber \\
 & =\expect[U]{f^{*}(\gamma_0(a,U)) \given A=a,W=\cdot}.\quad(\because\text{\cref{assum:deterministic-confounding}})\label{eq:fstar-intermediate}
\end{align}

From Assumption~\ref{assum:completeness-confounder},
\begin{align*}
    \expect{l(U) \mid A=a, W=w}=0\quad \forall w \in \mathcal{W} \quad\Leftrightarrow \quad l(u) = 0 \quad P_{U}\text{-}\mathrm{a.e.}
\end{align*}
for all functions $l \in L^2(P_{U|A=a})$. 
Hence, \eqref{eq:fstar-intermediate} and Assumption~\ref{assum:completeness-confounder}
implies 
\[
f^{*}(\gamma_0(a,U))=0 \quad P_{U}\text{-}\mathrm{a.s.}
\]
Then, the inner product between $f^{*}$ and $I(y)$
is given as follows:
\begin{align*}
\inner{f^{*}}{I}_{L^{2}(P_{Y|A=a})} & =\expect[Y|A=a]{f^{*}(Y)Y}\\
 & =\expect[U|A=a]{\expect[Y|U,A=a]{f^{*}(Y)Y}}\\
 & \overset{(c)}{=}\expect[U|A=a]{f^{*}(\gamma_0(a,U))\gamma_0(a,U)}\\
 & =0.
\end{align*}
Here, (c) uses the fact that $Y$ is deterministic given $(a,u)$. Hence, we have 
\[
I \in N(E_{a}^{*})^{\perp}.
\]
\end{proof}
\end{lem}

Now, we are able to apply Proposition \ref{prop:picard}, leading to the \cref{thm:main-exitence}.
\begin{proof}[Proof of \cref{thm:main-exitence}]
By Lemma \ref{lem:null-space-assumption}, the identity 
$I(y)$ is in $N(E_{a}^{*})^{\perp}.$ Therefore,
by Proposition \ref{prop:picard}, under the given assumptions, there exists a solution to \eqref{eq:inteq}. Letting the
solution be $h^*_{a}$ completes the proof. 
\end{proof}

\subsection{Identifiability in deterministic confounding} \label{sec:identifiability-in-deterministic-confounding}
Here, we show that the bridge function $h_0$ can be used to compute the structural function.


\begin{proof}[Proof of \cref{thm:main}]
First, we show that under \cref{assum:deterministic-confounding}, we have 
\begin{align}
    \expect{\gamma_0(\tilde a,U)|A=\tilde a, Y=y} = y. \label{eq:cond}
\end{align}
Let $B(\tilde a,y) = \{u\in\calU |\gamma(\tilde a,u) = y \}$. Then, we have 
\begin{align*}
    \expect{\gamma_0(\tilde a,U)|A=\tilde a, Y=y} &= \int \gamma_0(\tilde a,U) \intd P(U|A=\tilde a, Y=y)\\
    &= \int_{\neg B(\tilde a,y)} \gamma_0(\tilde a,U) \intd P(U|A=\tilde a, Y=y) + \int_{B(\tilde a,y)} \gamma_0(a,U) \intd P(U|A=\tilde a, Y=y)\\
    &= \int_{\neg B(\tilde a,y)} \gamma_0(\tilde a,U) \intd P(U|A=\tilde a, Y=y) + y \int_{B(\tilde a,y)} \intd P(U|A=\tilde a, Y=y)
\end{align*}
Now, from Bayes' theorem,
\begin{align*}
     P(U|A=\tilde a, Y=y) \propto P(Y=y|U,A=\tilde a)P(U|A=\tilde a).
\end{align*}
Since $Y$ is generated deterministically, the support of $P(Y=y|U,A=\tilde a)$ is $B(\tilde a,y)$. Therefore,
\begin{align*}
    \int_{\neg B(\tilde a,y)} \gamma_0(\tilde a,U) \intd P(U|A=\tilde a, Y=y) = 0, \quad \int_{B(\tilde a,y)} \intd P(U|A=\tilde a, Y=y) = 1.
\end{align*}
Hence, we have \eqref{eq:cond}. Using this, we have 
\begin{align*}
    &\expect{h_0(\tilde a,W)|A=\tilde a, Y=y} - y = 0\\
    \Leftrightarrow\quad&\expect[U|A=\tilde a,Y=y]{\expect{h_0(\tilde a,W)|U}}-\expect[U|A=\tilde a,Y=y]{\gamma_0(\tilde a,U)}=0\quad(\because \text{\eqref{eq:cond}, \cref{assum:structural}})
\end{align*}
If we assume informative outcome assumption \eqref{eq:informative-outcome}, this means
\begin{align*}
    \expect{h_0(\tilde a,W)|U} = \gamma_0(\tilde a,U),
\end{align*}
and thus
\begin{align*}
    f_\struct(\tilde a) = \expect[U]{\gamma_0(\tilde a,U)} = \expect[U]{\expect{h_0(\tilde a,W)|U}} = \expect{h_0(\tilde a,W)}.
\end{align*}

We can draw the same conclusion if we instead assume $A \indepe W|Y^{(\tilde a)}$ as
\begin{align*}
    \expect{h(\tilde a,W)} &\overset{(a)}{=} \expect[Y^{(\tilde a)}]{\expect{h(\tilde a,W)|Y^{(\tilde a)}, A=\tilde a}}\\
    &\overset{(b)}{=} \int \expect{h(\tilde a,W)|Y=y, A=\tilde a} P(Y^{(\tilde a)}=y)\intd y\\
    &\overset{(c)}{=} \expect{Y^{(\tilde a)}},
\end{align*}
where we used $A \indepe W|Y^{(\tilde a)}$ in (a), ``consisitency'' $\expect{h(\tilde a,W)|Y^{(\tilde a)}=y, A=\tilde a}=\expect{h(\tilde a,W)|Y=y, A=\tilde a}$ in (b), and $\expect{h(\tilde a,W)|A=\tilde a, Y=y} = y$ in (c). 
\end{proof}

\subsection{Identifiability in stochastic counfounding} \label{sec:identifiability-in-stochastic-confounding}
In this section, we derive the error bound in the stochastic confounding.
\begin{proof}
The proof is similar to \cref{thm:main}. First, we show that if additive noise is bounded $|\varepsilon| < M$, we have 
\begin{align}
    |\expect{\gamma_0(\tilde a,U)|A=\tilde a, Y=y} - y| < M \quad \forall y \in \calY. \label{eq:cond-stochastic}
\end{align}
Let $B(\tilde a,y,\varepsilon) = \{u\in\calU \mid |\gamma(\tilde a,u) - y|\leq M\}$. Then, we have 
\begin{align*}
    \expect{\gamma_0(\tilde a,U)|A=\tilde a, Y=y} &= \int \gamma_0(\tilde a,U) \intd P(U|A=\tilde a, Y=y)\\
    &= \int_{\neg B(\tilde a,y)} \gamma_0(\tilde a,U) \intd P(U|A=\tilde a, Y=y)+ \int_{B(\tilde a,y)} \gamma_0(\tilde a,U) \intd P(U|A=\tilde a, Y=y)
\end{align*}
Now, from Bayes' theorem,
\begin{align*}
     P(U|A=\tilde a, Y=y) \propto P(Y=y|U,A=\tilde a)P(U|A=\tilde a).
\end{align*}
Since  $|\varepsilon| < M$, the support of $P(Y=y|U,A=\tilde a)$ is $B(\tilde a,y)$. Therefore,
\begin{align*}
    \int_{\neg B(\tilde a,y)} \gamma_0(\tilde a,U) \intd P(U|A=\tilde a, Y=y)= 0,\quad \int_{B(\tilde a,y)} \intd P(U|A=\tilde a, Y=y) = 1
\end{align*}
Furthermore, we have 
\begin{align*}
    \int_{B(\tilde a,y)} \gamma_0(\tilde a,U) \intd P(U|A=\tilde a, Y=y) &\geq \int_{B(\tilde a,y)} (y-M) \intd P(U|A=\tilde a, Y=y)\\
    &=y-M.
\end{align*}
Similarly, we can also show 
\begin{align*}
    \int_{B(\tilde a,y)} \gamma_0(\tilde a,U) \intd P(U|A=\tilde a, Y=y) \leq y+M.
\end{align*}
Hence, we have \eqref{eq:cond-stochastic}. Using this, we have 
\begin{align*}
    &\expect{h_0(\tilde a,W)|A=\tilde a, Y=y} = y\\
    \Leftrightarrow\quad&|\expect[U|A=\tilde a,Y=y]{\expect{h_0(\tilde a,W)|U}}-\expect[U|A=\tilde a,Y=y]{\gamma_0(\tilde a,U)}|\leq M\quad(\because \text{\eqref{eq:cond-stochastic}, \cref{assum:structural}})\\
    \Leftrightarrow\quad&|\expect{h_0(\tilde a,W)|U}-\gamma_0(\tilde a,U)|\leq M\Xi \quad(\because \text{\eqref{eq:dens-ratio}})
\end{align*}
Hence, if we solve equation \eqref{eq:bridge-function}, we have 
\begin{align*}
    f_\struct(\tilde a) &= \expect[U]{\gamma_0(\tilde a,U)}\\
    &\leq  \expect[U]{\expect{h_0(\tilde a,W)|U} + M\Xi}\\
    &= \expect[W]{h_0(\tilde a,W)} + M\Xi.
\end{align*}
Similarly, we have $f_\struct(\tilde a) \geq \expect[W]{h_0(\tilde a,W)} - M\Xi$.
\end{proof}

\section{Derivation}\label{sec:derivation}
\paragraph{Two-stage Regression} Here, we present the proof of \cref{thm:two-stage-sol}. First, we recall the empirical estimate of conditional mean embedding as follows.

\begin{prop}[{[\citealp[Theorem 5]{Song2009Hilbert}]}]
    Given stage 1 samples $\{w_i, a_i, y_i\}$, the solution of stage 1 regression is given as 
    \begin{align*}
        \hat{\mu}_{W|A,Y}(a, y) = \sum_{i=1}^n \beta_i(a,y) \phi(w_i), \quad \vec{\beta}(a,y) = (K_{AA}\otimes K_{YY} + n\lambda I)^{-1}(\vec{k}_A(a)\odot \vec{k}_Y(y))
    \end{align*}
    where $\vec{k}_\calA(a) = (k(a_i, a))_i \in \mathbb{R}^n,\vec{k}_\calY(y) = (k(y_i, y))_i$.
\end{prop}

Given this, we consider stage 2 regression
\begin{align*}
    \hat{h} = \argmin_{h\in\calH_{\calA\calW}} \frac1n \sum_{i=1}^n \left(y_i - \braket[\calH_{\calA\calW}]{h, \phi_\calA(a_i) \otimes \hat{\mu}_{W|A,Y}(a_i, y_i)}\right)^2 + \eta\|h\|^2_{\calH_{\calA\calW}}
\end{align*}
can be seen as the kernel ridge regression with the kernel function $\tilde k$ is given as 
\begin{align*}
    \tilde k((a,y), (a',y')) &= \braket[\calH_{\calA\calW}]{\phi_\calA(a) \otimes \hat{\mu}_{W|A,Y}(a, y), \phi_\calA(a') \otimes \hat{\mu}_{W|A,Y}(a', y')}\\
    &= k_\calA(a,a') \left(\vec{\beta}^\top(a,y) K_{WW} \vec{\beta}(a',y')\right).
\end{align*}
Using the closed-form solution of kernel ridge regression (See \citep[Chapter 14.4.3]{murphy2013machine}) yields \cref{thm:two-stage-sol}.

\paragraph{Equivalence to \citetMastouri}
\citetMastouri~considered a different way of deriving the closed-form of the two-stage regression.

\begin{prop}[{\citetMastouri[Proposition 2]}]\label{prop:alt-two-stage}
    Given stage 1 samples $\{w_i, a_i, y_i\}_{i=1}^n$ and stage 2 samples  $\{\dot{a}_i, \dot{y}_i\}_{i=1}^m$, and regularizing parameter $(\lambda, \eta)$, we have 
    \begin{align*}
        \hat{h}(a,w) = \vec{k}_W^\top(w) \Gamma \vec{k}_{\dot{A}}(a),\quad \Gamma \in \mathbb{R}^{n\times m}, \quad \mathrm{vec}(\Gamma) = (B \bar\otimes I)(M + m\eta I)^{-1}\dot{\vec{y}} \in \mathbb{R}^{nm}
    \end{align*}
    where $\bar\otimes$ is the tensor product of associated columns of matrices with the same number of columns, and $\mathrm{vec}$ is row-wise vectorization as $\mathrm{vec}(\begin{bmatrix}
        a & b\\
        c & d
    \end{bmatrix}) = [a, b, c, d]^\top$. The definitions of $\vec{k}_W, \vec{k}_{\dot A}, B, M$ can be found in \cref{thm:two-stage-sol}.
\end{prop}

Although this learns $mn$ parameters, we show this solution is equivalent to \cref{thm:two-stage-sol} restated as follows.
\begin{align*}
    \hat{h}(a,w) = \vec{\alpha}^\top (\vec{k}_{\dot A}(a) \odot B^\top\vec{k}_W(w)), \quad \vec{\alpha} = (M + m\eta I)^{-1}\dot{\vec{y}} \in \mathbb{R}^m
\end{align*}
From the definition of $\bar\otimes$ and $\vec{\alpha}$, we have 
\begin{align*}
    \mathrm{vec}(\Gamma) = \begin{bmatrix}
        B_{11} &  & & \\
         &  B_{12}& & \\
        &&\ddots&\\
        &&&B_{1n}\\
         B_{21} &  & & \\
         &  B_{22}& & \\
        &&\ddots&\\
        &&&B_{2n}\\
        &&\hspace{-30pt}\vdots&\\
         B_{m1} &  & & \\
         &  B_{m2}& & \\
        &&\ddots&\\
        &&&B_{mn}\\
    \end{bmatrix} \vec{\alpha}.
\end{align*}
Hence, from definition of $\mathrm{vec}$, we have 
\begin{align*}
    \Gamma = \begin{bmatrix}
        (\vec{B}_{1\cdot} \odot \vec{\alpha})^\top\\
        (\vec{B}_{2\cdot} \odot \vec{\alpha})^\top\\
        \vdots\\
         (\vec{B}_{n\cdot} \odot \vec{\alpha})^\top
    \end{bmatrix} \in \mathbb{R}^{n\times m},
\end{align*}
where $\vec{B}_{i\cdot}$ is the $i$-th row vector of matrix $B$. Therefore, \cref{prop:alt-two-stage} can be written as 
\begin{align*}
    \hat{h}(a,w) = \sum_{i=1}^m k_W(w_i, w) (\vec{B}_{i\cdot} \odot \vec{\alpha})^\top \vec{k}_{\dot A}.
\end{align*}
From linearity, we have 
\begin{align*}
    \hat{h}(a,w) &= \sum_{i=1}^m k_W(w_i, w) (\vec{B}_{i\cdot} \odot \vec{\alpha})^\top \vec{k}_{\dot A}\\
    &=  \left(\sum_{i=1}^m  \vec{B}_{i\cdot}k_W(w_i, w) \odot \vec{\alpha}\right)^\top \vec{k}_{\dot A}\\
    &=  (B^\top \vec{k}_W(w) \odot \vec{\alpha})^\top \vec{k}_{\dot A}\\
    &=  (B^\top \vec{k}_W(w) \odot  \vec{k}_{\dot A})^\top\vec{\alpha}
\end{align*}
The last equality holds for all vectors $(\vec{a}\odot\vec{b})^\top\vec{c} = (\vec{a}\odot\vec{c})^\top\vec{b} = \sum_i a_ib_ic_i$. Hence, we showed \cref{prop:alt-two-stage} is equivalent to \cref{thm:two-stage-sol}.

\paragraph{Maximum Moment Restriction} Here, we present the proof of \cref{thm:mmr-sol}. Let $\vec{h} \in \mathbb{R}^n = (h(a_i,w_i))_i \in \mathbb{R}^{n}$. Then, we can rewrite empirical loss $\hatmmrloss$ as 
\begin{align*}
    \hat{h} &= \argmin_h \frac1{n^2} \vec{h}^\top G\vec{h} - \frac{2}{n^2} \vec{y}^\top G\vec{h} + \frac1{n^2}\vec{y}^\top G\vec{y} + \eta \|h\|_{\calH_{\calA\calW}}\\
    &= \argmin_h \frac1n (\sqrt{G/n} \vec{y} - \sqrt{G/n} \vec{h})^\top (\sqrt{G/n} \vec{y} - \sqrt{G/n} \vec{h}) + \eta \|h\|_{\calH_{\calA\calW}}\\
    &= \argmin_h \frac1n \sum_{i=1}^n ((\sqrt{G/n} \vec{y})_i - (\sqrt{G/n} \vec{h})_i)^2 + \eta \|h\|_{\calH_{\calA\calW}}.
\end{align*}
Here, we denote $(\vec{x})_i$ as the $i$-th element of vector $\vec{x}$. From reproducing characteristic, we have 
\begin{align*}
    (\sqrt{G/n} \vec{h})_i = \braket{\tilde\phi_i, h}, \quad \tilde\phi_i = \sum_{j=1}^n (\sqrt{G/n})_{ij}\phi_{\calA\calW}(a_j, w_j),
\end{align*}
where $(\sqrt{G/n})_{ij}$ is the $(i,j)$-element of $\sqrt{G/n}$. Hence, we can regard this as kernel ridge regression, where the target is $(\sqrt{G/n} \vec{y})_i$ and the feature is $\tilde\phi_i$. Since
\begin{align*}
    \braket[\calH_{\calA\calW}]{\tilde\phi_i, \tilde\phi_j} = (\sqrt{G/n} L \sqrt{G/n})_{ij},
\end{align*}
we have the closed-form solution as $\hat{h} = \vec{\beta}^\top \tilde{\vec{k}}(a,w)$
\begin{align*}
\vec{\beta} &= (\sqrt{G/n} L \sqrt{G/n} + n\eta I)^{-1} \sqrt{G/n}\vec{y}\\
&= \sqrt{n}(\sqrt{G} L \sqrt{G} + n^2\eta I)^{-1} \sqrt{G}\vec{y}\\
\tilde{\vec{k}}(a,w) &=\begin{bmatrix}
    \braket[\calH_{\calA\calW}]{\tilde\psi_1, \phi_{\calA\calW}(a, w)}\\
    \braket[\calH_{\calA\calW}]{\tilde\psi_2, \phi_{\calA\calW}(a, w)}\\
    \vdots\\
    \braket[\calH_{\calA\calW}]{\tilde\psi_n, \phi_{\calA\calW}(a, w)}
\end{bmatrix}\\
&= \sqrt{G/n}\vec{k}(a,w)
\end{align*}
This yields the closed-form solution in \cref{thm:mmr-sol}.

\section{Consistency} \label{sec:consistent-detail}
We derive the consistency in this section. The proof is mostly adopted from \citet{Rahul2020KernelProxy} and \citet{Mastouri2021Proximal}.

\paragraph{Two-stage Regression}  To derive the consistency of two-stage regression, we make the following assumptions.
\begin{assum} \label{assum:bridge-smooth}
    Let operator $C_{\mu}$ be $C_{\mu} = \expect{\phi_{\mathrm{2}}(A,Y) \otimes \phi_{\mathrm{2}}(A,Y)}$
    for $\phi_{\mathrm{2}}(A, Y) = \phi_\calA(A) \otimes \mu_{W|A,Y}(A,Y) \in \calH_{\calA\calW}$ and its eigen-decomposition be $\{\eta_i, \phi_i\}$.  We assume $\eta_i$, the $i$-th eigenvalue, can be bounded by $\eta_i \leq C i^{-b_0}$ for constant $C, b_0$. Furthermore, we assume that there exists $g\in\mathcal{H}_{\calA\calW} $ such that $h_0 = C_\mu^{\frac{c_0-1}2} g$ for some $c_0 \in (1,2]$.
\end{assum}
This assumption is made in the analysis of various kernel-based methods \citep{Rahul2020KernelProxy,Mastouri2021Proximal,Fischer2020Sobolev}. The constant $b_0$ is known as the effective dimension of $\calH_{\calA\calW}$, and $b_0 = \infty$ when RKHS $\calH_{\calA\calW}$ is of finite dimension.  The constant $c_0$ measures the smoothness of the bridge function: a larger $c_0$ means a smoother bridge function. We assume a similar condition on operator $C_{W|A,Y}$.
\begin{assum} \label{assum:operator-smooth}
    Let the covariance operator of $\calH_{\calA\calY}$ be $C_{(A,Y),(A,Y)} = \expect{\phi_{\calA\calY}(A, Y)\otimes\phi_{\calA\calY}(A, Y)}$. We assume $\eta_i$, the $i$-th eigenvalue of $C_{(A,Y),(A,Y)}$, can be bounded by $\eta_i \leq C i^{-b_1}$ for constant $C, b_1$. Furthermore, we assume that there exists $G \in \calH_{\calW(\calA\calY)}$ that $C_{W|A,Y} = G \circ C^{\frac{c_1-1}2}_{(A,Y),(A,Y)}$ for some $c_1 \in (1,2]$.
\end{assum}

Here, we denote ${}^*$ as the adjoint of the operator. Given these assumptions, we obtain the following consistency result.
\begin{prop}\label{prop:2sr-consistency}
    Assume \cref{assum:bridge-smooth,assum:operator-smooth}. Given stage 1 samples $\{w_i, a_i, y_i\}_{i=1}^n$ and stage 2 samples  $\{\dot{a}_i, \dot{y}_i\}_{i=1}^m$, and let $\lambda = n^{-\frac{1}{c_0+1/b_0}}$ and $n = m^{\xi \frac{c_0+1/b_0}{c_0-1}}$ for some $\xi>0$. Then,
    \begin{align*}
        \sup_{a,w} |h_0(a,w) - \hat{h}(a,w)| = O_P(m^{-\frac{\xi c_1-1}{2(c_1+1)}}),
    \end{align*}
    with $\eta = m^{-\frac{\xi}{c_1+3}}$ when $\xi \leq  \frac{c_1 + 3}{c_1 + 1/b_1}$. Otherwise, we have 
    \begin{align*}
        \sup_{a,w} |h_0(a,w) - \hat{h}(a,w)| =  O_P(m^{-\frac{c_1-1}{2(c_1+1/b_1)}}),
    \end{align*}
    with $\eta = m^{-\frac{1}{c_1+1/{b_1}}}$. These upper bounds can be translated to the estimation error of structural function as 
    \begin{align*}
        &|f_\struct(\tilde a) - \hat{f}_\struct(\tilde a) | \leq \sup_{w} |h_0(\tilde a,w) - \hat{h}(\tilde a,w)| + O_P(n^{-\frac12}),
    \end{align*}
    under the condition of \cref{thm:main}.
\end{prop}
The proof is a direct application of Theorem 3 in \citet{Rahul2020KernelProxy}. From this, we can see that SKPV can consistently recover the bridge function and structural function. As in \citet{Rahul2020KernelProxy}, we observe that the best data ratio is $n = m^{\frac{c_1+3}{c_1+1/b_1}\frac{c_0+1/b_0}{c_0-1}}$ from which $n \gg m$. The best rate for bridge function estimation under this data size ratio is $O_P(m^{-\frac{c_1-1}{2(c_1+1/b_1)}})$ which is the same as for kernel ridge regression \citep{Fischer2020Sobolev}.

\paragraph{Maximum Moment Restriction} For maximum moment restriction approach, we use the same assumption as \citetMastouri[Assumption 16].
\begin{assum} \label{assum:mmr-smooth}
    Let $\{\eta_i, \phi_i, \psi_i\}$ be a singular decomposition of operator $T = \expect{\phi_{\calA\calW}(A,W) \otimes \phi_{\calA\calY}(A,Y)}$. We assume $h_0$ is in a $\gamma$-regularity space $\calH^{\gamma}_{\calA\calW} $ that is defined as 
    \begin{align*}
        \calH^{\gamma}_{\calA\calW} = \left\{f \in \calH_{\calA\calW} ~\middle|~ \sum_{i=1}^\infty \frac{\braket[\calH_{\calA\calW}]{f, \phi_i}^2}{\eta_i^{2\gamma}}\right\}
    \end{align*}
\end{assum}
This assumption is similar to \cref{assum:bridge-smooth}: a larger $\gamma$ means a smoother bridge function. Given this, we can show the consistency of the SPMMR method.
\begin{prop}\label{prop:mmr-consistency}
    Given \cref{assum:mmr-smooth} and samples $\{w_i, a_i, y_i\}$ size of $n$. Then, we have 
    \begin{align*}
        \|\hat{h} - h_0\|_{\calH_{\calA\calW}} \leq O\left(n^{-\frac12 \min\left(\frac12, \frac{\gamma}{\gamma+2}\right)}\right)
    \end{align*}
    with $\lambda = n^{-\frac12 \min\left(\frac12, \frac{2}{\gamma+2}\right)}$.
\end{prop}

The proof is the direct application of \citetMastouri[Theorem 3]. From this, we can see that the best rate $O(n^{-1/4})$ is achieved when $\gamma \geq 2$. Again, this best rate is the same as kernel ridge regression or SKPV. Furthermore, \citetMastouri~discussed the relation of the two-stage regression approach and maximum moment restriction approach in the asymptotic case; this relation also applies to our SKPV and SPMMR.

\section{Experiment Details}\label{sec:exp-detail}
In this section, we provide the details of experiments with the regularization parameter tuning. All experiments are run on MacBook Air within a few minutes.

\paragraph{Regularization Parameter Tuning for SKPV and SPMMR}
We follow the procedure in \citetMastouri to tune the regularization parameter $(\lambda, \eta)$ for SKPV and $\lambda$ in SPMMR. In SKPV, to select the best $\lambda$ in stage 1 regression, we use leave-one-out error for ${\hat{\mathcal{L}}}^{\mathrm{cond}}$. This can be computed as the closed-form solution
\begin{align*}
    {\hat{\mathcal{L}}}^{\mathrm{cond-loo}}(\lambda) = \mathrm{tr}({\tilde H_1}^{-1}H_1 K_{WW} H_1{\tilde H_1}^{-1})
\end{align*}
where $H_1 = I - (K_{AA}\odot K_{YY})(K_{AA}\odot K_{YY} + n\lambda I)^{-1},$
and $\tilde H_1$ is the diagonal matrix with the same diagonals $H_1$. Given the best parameter $\lambda = \argmin  {\hat{\mathcal{L}}}^{\mathrm{cond-loo}}$, we can tune the second stage regularizer $\eta$ by using the stage 1 data as ``validation data'' in stage 2 loss:
\begin{align*}
    \eta = \argmin \frac1n \sum_{i=1}^n \left(y_i - \braket[\calH_{\calA\calW}]{\hat{h}, \phi_\calA(a_i) \otimes \hat{C}_{W|A,Y}\phi_{\calA\calY}(a_i, y_i)}\right)^2,
\end{align*}
where $\hat{h}$ is the solution in \cref{thm:two-stage-sol}. In SPMMR, we cannot use leave-one-out error to tune $\lambda$ in SPMMR since $\hatmmrloss$ is a V-statistic. We selected the best  $\lambda$ by minimizing the empirical loss $\hatmmrloss$ over a held-out validation set.

\begin{figure}
    \centering
    \includegraphics[width = 0.6\textwidth]{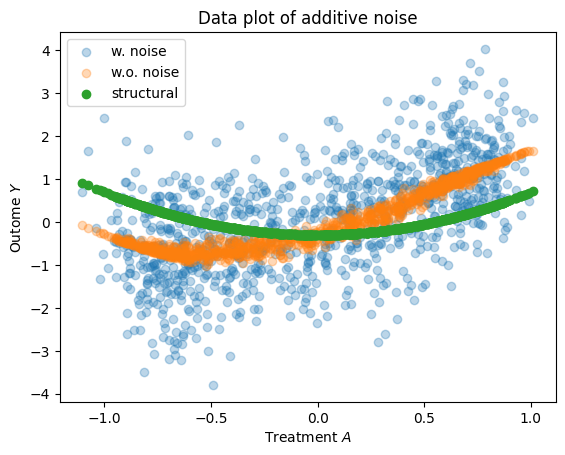}
    \label{fig:data_plot}
    \caption{The plot of data samples with additive noise (blue) and without additive noise (orange). Green curve is true structural function $Y = f_\struct(A)$}
\end{figure}

\paragraph{Additive Noise Increases the Support}

In the sensitivity analysis on the deterministic assumption, we observe that adding Gaussian noise $\mathcal{N}(0, 1)$ can increase the performance. To explore this, we plot the data samples of treatment $A$ and outcome $Y$ in \cref{fig:data_plot}. From it, we can see that the true structural function is completely away from the support when noise is zero. However, it is in the well-supported domain when we add the Gaussian noise.  

\end{document}